\documentclass[journal]{IEEEtran}
\usepackage{tikz}
\usepackage{pgfplots}
\usetikzlibrary{patterns}
\pgfplotsset{compat=1.16}
\usepackage{wrapfig}
\usepackage{subcaption}
\usepackage{tikz}
\usepackage{pgfplots}
\usetikzlibrary{patterns}
\pgfplotsset{compat=1.16}
\usepackage{wrapfig}
\usepackage{subcaption}
\usepackage{smartdiagram}  
\usepackage[edges]{forest}
\usetikzlibrary{arrows.meta}
\usepackage{rotating}
\usetikzlibrary{shapes,arrows,positioning}

\usepackage{cite}
\usepackage[utf8]{inputenc}
\usepackage{blindtext}
\usepackage{tabularx}
\usepackage{color}
\usepackage{booktabs}
\usepackage{times}
\usepackage{soul}
\usepackage{url}
\usepackage[hidelinks]{hyperref}
\usepackage{amsmath}
\usepackage{booktabs}
\usepackage{url}
\usepackage{color}
\usepackage{booktabs}
\usepackage{graphicx}
\usepackage{xspace}
\usepackage{xcolor}
\usepackage{soul}
\usepackage{fancyhdr}
\usepackage{graphicx}
\usepackage{graphics}
\usepackage{soul}
\usepackage{url}
\usepackage[hidelinks]{hyperref}
\usepackage{graphicx}
\usepackage{amsmath}
\usepackage{booktabs}
\usepackage{booktabs}
\usepackage{graphicx}
\usepackage{xspace}
\usepackage{soul}
\usepackage{fancyhdr}
\usepackage{csquotes}
\usepackage{tikz}
\usepackage{pgfplots}
\pgfplotsset{compat=1.16}
\usepackage{wrapfig}
\usepackage{subcaption}
\usepackage{csquotes}
\usepackage{enumitem}
\usepackage{amssymb}
\usepackage{amsmath}
\usepackage{multirow}
\usepackage{graphicx}
\usepackage{multirow}
\usepackage{placeins}

\usepackage[T1]{fontenc}
\usepackage{algorithm}
\usepackage{algpseudocode}
\usepackage{stfloats} 
\usepackage{pgfplots}
\usepackage{xcolor}
\usepackage{float}
\usepackage{silence}

\usepackage{caption}

\usepackage{subcaption}
\begin{document}

\title{Pseudo-Labeling Driven Refinement of Benchmark Object Detection Datasets via \textcolor{black}{Analysis of Learning Patterns}}

\author{Min~Je~Kim, Muhammad Munsif,~\IEEEmembership{Student Member,~IEEE}, Altaf Hussain,~\IEEEmembership{Student Member,~IEEE}, Hikmat Yar,~\IEEEmembership{Student Member,~IEEE},
              Sung Wook~Baik,~\IEEEmembership{Senior Member,~IEEE}

\thanks{This work was supported by a National Research Foundation of Korea (NRF) grant funded by the Korea government (MSIT), Grant/Award Number: (2023R1A2C1005788). (Corresponding author: Sung Wook Baik)
\newline
\indent Min Je Kim, Muhammad Munsif, Altaf Hussain, Hikmat
Yar, and Sung Wook Baik are with Sejong University, Seoul 143-747, South Korea. (Email: \href{mailto:}{minjekim3797@gmail.com}; \href{mailto:}{munsif@ieee.org}; \href{mailto:}{a.hussain@ieee.org};  \href{mailto:}{hikmat@ieee.org}; 
\href{mailto:}{sbaik3797p@gmail.com})

}

}

\maketitle

\begin{abstract} 
Benchmark object detection (OD) datasets play a pivotal role in advancing computer vision applications such as autonomous driving, robotics, and surveillance, as well as in training and evaluating deep learning-based state-of-the-art detection models. Among them, MS-COCO has become a standard benchmark due to its diverse object categories and complex scenes. However, despite its wide adoption, MS-COCO suffers from various annotation issues, including missing labels, incorrect class assignments, inaccurate bounding boxes, duplicate labels, and group labeling inconsistencies. These errors not only hinder model training but also degrade the reliability and generalization of OD models. To address these challenges, we propose a comprehensive refinement framework and present MJ-COCO, a newly re-annotated version of MS-COCO. Our approach begins with loss and gradient-based error detection to identify potentially mislabeled or hard-to-learn samples. Next, we apply a four-stage pseudo-labeling refinement process: (1) bounding box generation using invertible transformations, (2) IoU-based duplicate removal and confidence merging, (3) class consistency verification via expert objects recognizer, and (4) spatial adjustment based on object region activation map analysis. This integrated pipeline enables scalable and accurate correction of annotation errors without manual re-labeling. Extensive experiments were conducted using one-stage (RetinaNet, YOLOv3, YOLOX) and two-stage (Faster R-CNN, Libra R-CNN) OD models across four validation datasets: MS-COCO, Sama COCO, Objects365, and PASCAL VOC. 
Models trained on MJ-COCO consistently outperformed those trained on MS-COCO, especially on high-quality validation sets, achieving improvements in Average Precision (AP) and AP$_S$ metrics.
MJ-COCO also demonstrated significant gains in annotation coverage: for example, the number of small object annotations 
\textcolor{black}{increased by more than 200,000}
compared to MS-COCO. These results confirm that MJ-COCO offers a more accurate, robust, and scalable alternative for modern OD tasks, improving both model performance and dataset reliability. Our MJ-COCO dataset's annotations are publicly available to the research community at \url{https://www.kaggle.com/datasets/mjcoco2025/mj-coco-2025}.

\end{abstract}

\IEEEpeerreviewmaketitle

\section{Introduction}
 \label{sec:Introduction}

Object Detection (OD) is a fundamental task in Computer Vision (CV) that involves identifying and localizing instances of objects, such as animals, humans, vehicles, etc., within images. 
\textcolor{black}{It serves as a crucial foundation for numerous real-world applications, including robotics, healthcare, self-driving cars, and satellite and aerial imagery analysis\cite{bolya2020tide}.}
Its primary objective is to develop computational models capable of answering a key question for vision-based applications:
"What objects are present, and where are they located". The recent development in Deep Learning (DL) models has significantly advanced OD, making it a major focus of research. However, the effectiveness of DL models heavily depends on the availability and quality of OD benchmark datasets. High-quality OD datasets play a critical role in training robust models by providing diverse, well-annotated images that improve generalization and performance across different scenarios. 
\begin{figure}[t]
    \centering
    \includegraphics[width=1.01\linewidth]{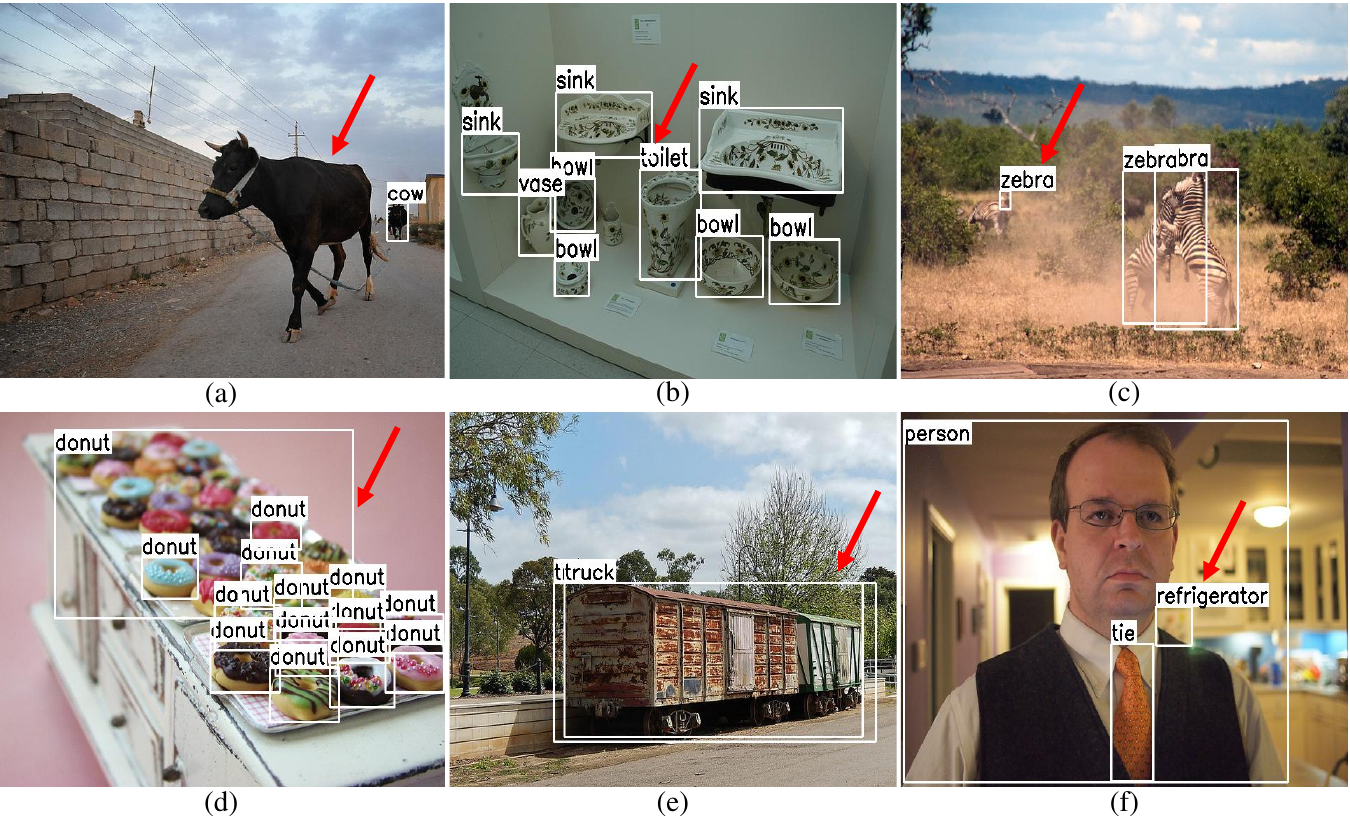}
    \caption{ Visual representation of annotation errors in the MS-COCO dataset for OD. (a) Missing object labels, (b) Incorrect class assignments, (c) Inaccurate bounding boxes, (d) Group labeling, (e) Duplicate labeling, and (f) Incorrect object estimations.}
    \label{COCO_Errors}
\end{figure}
Early efforts in CV primarily focused on image classification, but with the increasing demand for precise object localization and segmentation, the need for dedicated OD datasets became apparent. Therefore, several OD datasets have been developed to achieve precise detection and localization of targeted objects. One of the earliest datasets for OD was developed in 1998 by the Massachusetts Institute of Technology~\cite{papageorgiou1998trainable}.

Navneet Dalal and Bill Triggs introduced the INRIA Person dataset \cite{dalal2005histograms}, However, the dataset has some limitations, including a limited number of annotated images and categories, missing annotations for some humans and the lack of labels for different objects and complex scenarios [4].
\textcolor{black}{With the emergence of object OD datasets, a pioneering contribution to both OD and classification was made by the PASCAL VOC dataset in 2005 \cite{everingham20062005}}. However, it had several limitations, such as the limited number of classes, images, and corresponding bounding boxes, which restricted its applicability to a broader range of OD tasks.~\textcolor{black}{To address these limitations, the PASCAL VOC challenge released updated versions of the dataset each year from 2006 to 2012. Among these, the PASCAL VOC 2012 dataset~\cite{everingham2011pascal} is the most widely used dataset in OD research, as it includes} 20 object classes, covering a more diverse range of categories such as animals, vehicles, and indoor objects. However, it falls short in representing complex scenes with high object density, diverse occlusions, and contextual relationships, which limits its applicability to state-of-the-art OD tasks. In 2014, Microsoft introduced the Microsoft Common Objects in Context (MS-COCO) dataset, a new benchmark for OD in CV. It contains 164,000  annotated images with 80 object categories, providing detailed bounding boxes per instance, considered as a valuable dataset in the field. Despite its advancements, MS-COCO-2014 has some limitations, including annotation inconsistencies, limited dataset size for training complex models, and class imbalances that can affect model training and evaluation. Additionally, the dataset requires more images to enhance model performance, as its size may still be insufficient for training complex DL models. To address the limitations of the previous version, Microsoft released an updated and refined version of MS-COCO in 2017 \cite{mscoco}. This extended version aimed to strengthen its role as a benchmark in CV research by improving annotation quality and increasing the number of images, making it a more reliable and comprehensive dataset for training and evaluating modern OD models. 

\subsection{Research Gap}
Despite significant efforts in developing OD datasets, each comes with inherent challenges that limit its applicability to real-world scenarios. The MS-COCO-2017 dataset introduces 80 object categories, a larger, and more diverse image dataset. Nevertheless, early versions like MS-COCO-2014 had annotation inconsistencies and insufficient data volume, while the improved 2017 version still faces issues like missing object labels, incorrect class assignments, inaccurate bounding
boxes, group labeling, duplicate labeling, and incorrect object estimations, as shown in the Figure~\ref{COCO_Errors}. Collectively, these datasets illustrate a clear evolution toward richer and more realistic benchmarks, yet they underscore the continuing need for more correctly, comprehensively annotated, and context-aware datasets for modern OD models. 
To address the aforementioned challenges in the \textcolor{black}{MS-COCO-2017} dataset, researchers have explored various approaches to mitigate annotation errors. These efforts can be categorized into manual (human-based) solutions and semi-automatic methods (a combination of human and machine involvement) for error detection and correction. Manual efforts involve human-based re-annotations \cite{zimmermann2023benchmarking}, While this human-centric approach aims to enhance labeling accuracy, it inherently introduces challenges, such as annotation inconsistencies and potential human errors. Lastly, they ignore the smallest object in the re-annotation, which hinders the model for real-world applications.
Furthermore, for the correctness and consistency issues in bounding box annotations, Ma et al.~\cite{ma2022effect} proposed a data-centric AI method to improve annotation quality by employing well-trained human annotators and detailed guidelines for high-quality re-annotation. The re-annotation covers standards for crowding objects, limited visibility objects, and covers few object categories. Additionally, human-based re-annotation is time-consuming and labor-intensive, limiting its practicality for large-scale, evolving datasets. Some researchers used a semi-automatic approach, which significantly reduced the impact of annotation noise, providing a practical solution for real-world OD. While re-annotation using semi-automatic labeling techniques has been proposed to mitigate these challenges, ensuring high-quality annotations at scale remains a complex and labor-intensive task.

\subsection{Contributions}

\textcolor{black}{Although several methods have been proposed to correct annotation errors, inconsistencies persist and continue to negatively impact training and evaluation~\cite{zimmermann2023benchmarking}~\cite{ma2022effect}.}
To address these issues, OD demands more accurate and scalable annotation methods for dataset refinement. 
Therefore, we present a comprehensive framework for labeling error detection and pseudo-labeling. Firstly, a dataset is  prepared to \textcolor{black}{identify }
correct and incorrect labels to support model optimization.
It integrates the Faster R-CNN architecture and applies pseudo-labeling, object region detection, \textcolor{black}{and activation map analysis techniques to refine and correct label annotations, thereby enhancing overall dataset quality.}
The major contributions of the proposed framework are as follows:

\begin{itemize}
    \item We propose a novel automatic annotation error detection and correction framework for large-scale OD benchmark datasets. Our method employs a unified RPN for proposal generation, followed by unsupervised loss reconstruction to identify annotation errors. Furthermore, we introduce a spatial gradient-based activation map mechanism to support pseudo-labeling to improve the performance and robustness of the annotation correction pipeline and ultimately the OD performance.

     \item To address the prevalent issue of noisy annotations in large-scale datasets, we design a dynamic strategy combining Faster R-CNN with an \textcolor{black}{AutoEncoder}-based loss and gradient monitoring. This proposal generation and learning monitoring mechanism, enables precise and efficient identification of annotation anomalies without the need for manual supervision.
     
    \item For systematic re-annotation, we introduce a robust four-stage pseudo-labeling refinement framework aimed at producing semantically consistent and spatially accurate re-annotations. Unlike traditional pseudo-labeling methods, our approach leverages invertible transformation-based multi-angle consistency, IoU-driven redundancy filtering, hierarchical validation via expert objects recognizer, and spatial adjustment based on object region activation map analysis. The resulting re-labeled dataset, called MJ-COCO, demonstrates improved generalization and reliability across multiple detection backbones.

        \item We conduct comprehensive experiments on the MS-COCO benchmark and its variants, including Sama-COCO, Objects36, and PASCAL VOC. Our proposed framework is evaluated using both one-stage detectors (RetinaNet, YOLOv3, YOLOX) and two-stage detectors (Faster R-CNN, Libra R-CNN). The results consistently show enhanced detection performance and correction effectiveness, outperforming existing state-of-the-art methods.
        Additionally, qualitative evaluations confirm the practicality of our framework in real-world OD applications.

    \end{itemize}

Section II provides a comprehensive review of related work to establish the research context. Section III details the architecture and methodology of the proposed framework. Section IV presents extensive experimental results, including comparisons with state-of-the-art models to demonstrate the effectiveness of our approach. Finally, Section V summarizes the key findings and discusses potential directions for future research.
 \section{Related Work} \label{sec:relatedwork}
In this section, we present a comprehensive review of the limitations in existing OD datasets, highlighting the challenges associated with annotation error detection and examining current pseudo-labeling techniques used for correcting and enhancing re-annotation.
\subsection{Limitations in OD Datasets} 
OD datasets play a crucial role in model training and evaluation by providing bounding boxes and class labels that enable models to identify and locate objects within images. Before the advent of OD datasets, most datasets were primarily designed for image classification tasks, where models identified the entire image as a single class, lacking the ability to perform image localization or detect multiple objects within the same image. These limitations led to the development of benchmark datasets designed to both identify and localize objects. For instance, Papageorgiou et al.~\cite{papageorgiou1998trainable} introduced an OD dataset comprising 1,848 frontal and rear images, marking one of the earliest efforts in pedestrian detection. However, it focused exclusively on pedestrians and provided limited class diversity, restricting its usefulness in broader real-world applications. Navneet Dalal and Bill Triggs proposed the INRIA Person detection dataset \cite{dalal2005histograms}, consisting of 1,805 cropped images of humans. However, it had several limitations, including missing annotations for some individuals, a lack of complex environments, and a narrow focus solely on human detection. In 2005, Zisserman et al.~\cite{everingham20062005}, introduced the PASCAL VOC 2005 dataset, which played a significant role in the OD domain by offering four object classes and a standardized benchmark. However, its limited number of classes, small image count, and lack of detailed bounding box annotations restricted its applicability to diverse OD tasks. To cope with these limitations, several PASCAL VOC datasets were introduced in the subsequent years such as PASCAL VOC 2006 \cite{everingham2006pascal}, 2007 \cite{everingham2008pascal}, 2008 \cite{hoiem2009pascal}, 2009 \cite{everingham2009pascal}, 2010 \cite{everingham2010pascal}, 2011 \cite{everinghampascal}, and 2012 \cite{everingham2011pascal}. Among these datasets, the PASCAL VOC 2007 and 2012 is the most widely used datasets. The PASCAL VOC 2007 dataset includes 20 object classes across 9,963 images, covering animals, vehicles, and indoor objects. However, it struggled to capture real-world complexities such as varied lighting and occlusion. The PASCAL VOC 2012 improved annotation quality for better localization but suffered from class imbalance, favoring frequent objects over rare ones. Despite its growth, the dataset remained too small for effective DL model training. In 2014 and 2017, Microsoft released the MS-COCO-2014 and MS-COCO-2017 datasets, designed for OD, instance segmentation, and keypoint detection. These datasets includes 80 object classes captured from real-world environments, offering rich contextual information and diverse scenes. The MS-COCO-2014 dataset faced limitations such as annotation quality, class imbalance, and limited size \cite{oksuz2020imbalance} . To address these, Microsoft released MS-COCO-2017, which improved labeling precision and bounding box accuracy. However, due to manual annotation by multiple annotators, errors still persist. Some researchers have categorized these errors and employed DL-based OD algorithms to improve detection performance. In a research \cite{borjibreaking}, the author provided compressive analysis of MS-COCO dataset and highlighted several types of annotation error, which are categories as follows: (1) Missing annotations, where objects are present but unlabeled, often due to occlusion, small size, or cropping; (2) Incorrect labels, where nonexistent or misclassified objects lead to false positives; (3) Localization errors, bounding boxes that are misaligned or fail to capture the full object; (4) Duplicate annotations, where multiple boxes for the same object or one box covering multiple objects; (5) Inconsistent annotations, similar objects labeled differently across images, affecting generalization;
(6) Complex cases, challenging scenes with cluttered backgrounds or varied object sizes; and
(7) Ambiguous examples, objects difficult to recognize, leading to subjective or inconsistent labeling. To address the aforementioned issue Sama-COCO \cite{zimmermann2023benchmarking}, manually refine the dataset to improve the performance and annotation consistency. However, manual annotation remains error-prone, labor-intensive, and time-consuming. Furthermore, in \cite{tong2023rethinking}, the authors refined the MS-COCO dataset and proposed three novel subset of MS-COCO dataset, including Mini6K, Mini2022, and Mini6KClean. Mini6K consists of 6,000 images and 44,217 annotation across 80 classes. Despite this effort, the dataset still suffers from annotation issues such as labeling errors, missing annotations, and incorrect class assignments. To cope with these, they proposed another subset of the dataset named Mini2022, which consists of 729 images, 78 classes, and 23,391 manual re-annotated objects. Furthermore, \cite{ma2022effect} proposed A data-centric AI approach to improve annotation accuracy and consistency in OD datasets by leveraging skilled human annotators guided by comprehensive annotation guidelines. This approach also employed AI systems to automatically detect and correct annotation errors. Although this approach aimed to enhance the overall dataset quality, the experimental results showed that re-annotation on the MS-COCO dataset tended to decrease the mean Average Precision (mAP). Therefore, numerous research studies have focused on anomaly detection in datasets, proposing pseudo-labeling methods to improve OD accuracy.
\begin{figure*}[b]
    \centering
    \includegraphics[width=1.0\linewidth, trim=0cm 0.2cm 0cm 0.2cm, clip]{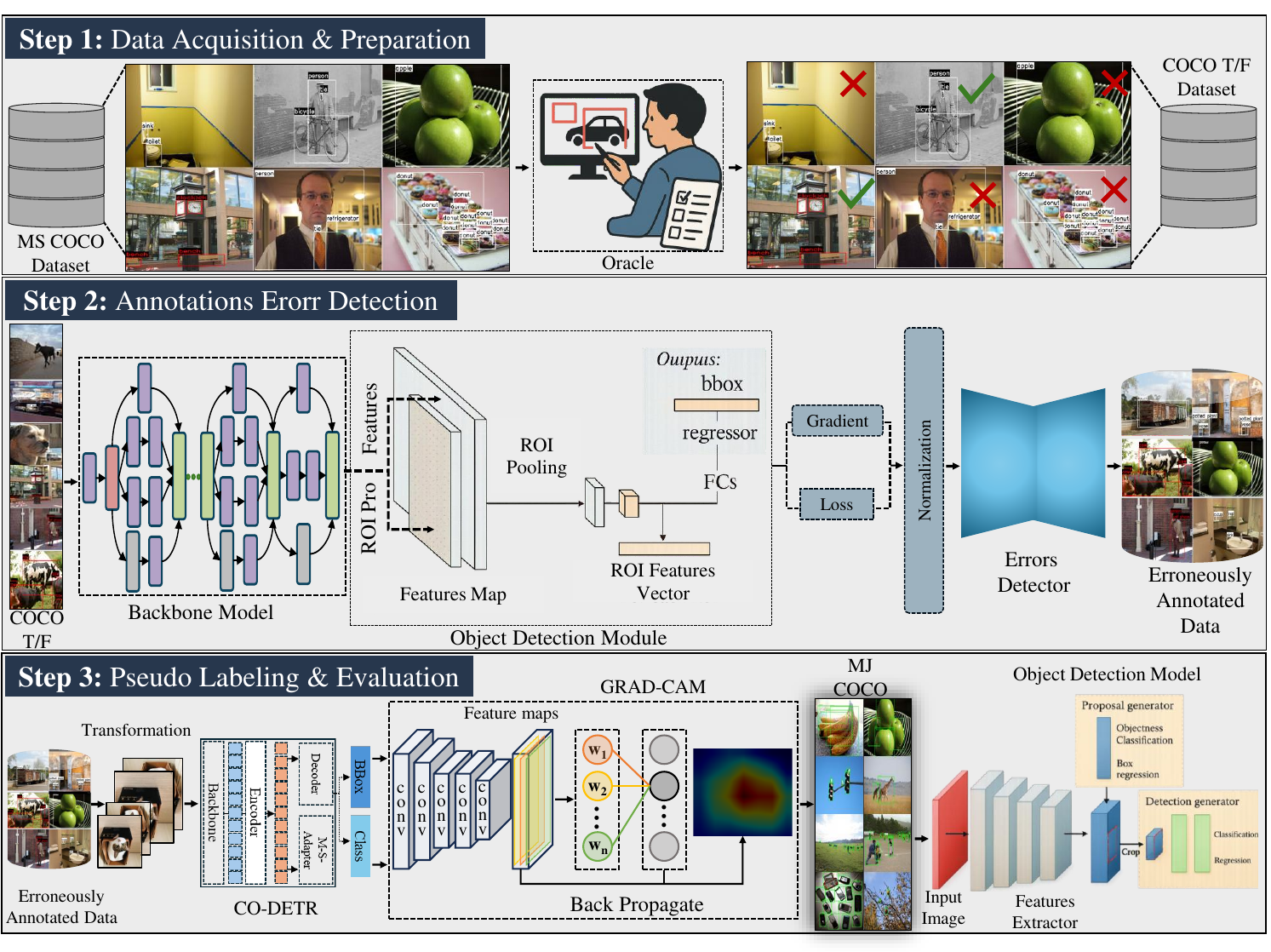}
    \caption{Overview of the proposed framework: Consist of three stages: data preparation, error detection using a hybrid two stage detector and \textcolor{black}{AutoEncoder} followed by pseudo labeling utilizing Grad-CAM with evaluation of object detectors on the MJ-COCO dataset.}
    \label{Framework}
\end{figure*}
\subsection{Annotation Error Detection and Pseudo Labeling in OD Datasets} 
The performance of OD models is highly influenced by the quality of training data, and annotation errors, which can significantly reduce the model’s generalization capability during training. Therefore, various researchers proposed different approaches for annotation error detection in OD datasets. For instance, Zhu et al.~\cite{zhu2015diagnosing} analyzed the impact of object size, color contrast, iconic view, aspect ratio, shape regularity and texture. Their experimental results showed for small, irregularly shaped, and objects with low color contrast, significantly reduced the model performance. Jeffri et al.~ \cite{murrugarra2022can}, experimentally demonstrated the bounding box annotations discrepancies, which significantly affect final OD evaluation metrics mAP being particularly sensitive in the case of small objects. Annotation errors in benchmark datasets can degrade detection performance. To address this, \cite{tkachenko2023objectlab} proposed ObjectLab to detect such errors in the \textcolor{black}{MS-COCO} dataset and experimentally showed that approximately 5\% of the images contained missing object annotations, 3\% had localization errors, and 0.7\% exhibited class assignment errors. In \cite{schubert2024identifying}, the author created a benchmark dataset with four label error types and proposed a multi-stage detection method using classification and regression losses. It outperformed baselines with low false detection rates and high reliability. 
However, annotation errors in OD datasets are still an unresolved issue, due to these limitations, recent studies have shifted focus from explicitly removing anomalous data to designing robust OD models capable of maintaining high performance even in noisy environments. For instance, Xu et al.~\cite{xu2021training}, proposed a Meta-Refine-Net to train object detectors from noisy category labels and imprecise bounding boxes. Firstly, they down-weighting mislabeled proposals to reduce classification loss, then refining imprecise bounding boxes to benefit regression, and lastly, joint learning category and localization for more accurate annotations. It is model-agnostic and effective with minimal clean data (<2\%), showing promising performance on PASCAL VOC 2012 and MS-COCO 2017. In another approach \cite{liu2022robust} presented an object-aware multiple instance learning approach, which considers each image as a collection of object annotations and handles label errors by identifying the most trustworthy annotations and expanding them selectively. Furthermore,  Li et al.~\cite{li2022pseco}. proposed a semi supervised OD method named PseCo, that enhance pseudo-labeling and consistency training. It addresses poor localization in pseudo boxes using prediction-guided label assignment and positive-proposal consistency voting. To improve scale invariance, it represents multi-view scale-invariant learning by aligning feature-level consistency across scales. PseCo outperforms the state-of-the-art Soft Teacher on \textcolor{black}{MS-COCO} by up to 2.0 points with only 1–10\% labeled data, while also cutting training time in half. In \cite{zhou2022dense} proposed a dense prediction which is the united and straightforward form of pseudo-label. Dense pseudo-label does not involve any post-processing, thus retaining richer information. The authors introduced a region selection approach to highlight the key information while suppressing the noise carried by dense labels and achieved better performance on MS-COCO and PASCAL VOC datasets. A  pseudo-label enhancement framework is proposed in \cite{yang2025pseudo}, which integrates Class Activation Maps Reassembly (CAMR) and a Composite Scoring Module (CSM). CAMR improves object localization by combining class priors from CAMs with the semantic grouping of self-supervised Vision Transformers (ViTs). It propagates localization cues to ViT feature maps and refines them using patch affinity and a multi-threshold strategy to generate more accurate pseudo bounding boxes. CSM replaces one-hot labels with soft-class labels, incorporating both localization precision and classification confidence. Evaluated on PASCAL VOC and MS-COCO, pseudo-label enhancement outperforms existing methods, achieving a 7.1\% mAP gain on PASCAL VOC 2007. Fernandez et al.~\cite{garcia2025enhancing}, improves OD performance by mining reliable pseudo-labels from unlabeled data, addressing performance gaps caused by limited annotations. It introduces class confirmation to filter misclassified labels using class prototypes and box confirmation to remove poorly localized boxes via IoU estimation and improve the performance over MS-COCO and PASCAL VOC. The aforementioned methods have shown promising results in enhancing model performance by refining annotation quality. However, many rely on human annotators, which is both time-consuming and costly. More recent approaches utilize automatic re-annotation through pseudo-labeling, but these often introduce noise, negatively impacting training efficiency and accuracy. Moreover, most pseudo-label filtering techniques focus solely on category correctness, neglecting the localization quality and other errors discussed earlier in this section. This imbalance limits their effectiveness, as high classification confidence does not necessarily imply precise object localization; both are critical for reliable OD. Therefore, it becomes imperative to develop a novel, highly accurate, and computationally efficient automatic method specifically designed for the reliable detection and effective correction of annotation errors, ensuring improved data quality and model performance.


\section{Methodology} \label{sec:proposedmethod}
This section presents a comprehensive overview of the proposed framework for labeling error detection and pseudo-labeling. The framework primarily consists of three stages, with the first stage focusing on data preparation to identify correct and incorrect labels for subsequent model optimization. It integrates Faster R-CNN \cite{ren2016faster} architectures with loss and gradient monitoring diffusion into a unified system to enhance the performance of automatic labeling error detection. For the pseudo-labeling process, object region detection and Grad-CAM-based techniques are employed to refine and correct label annotations. This approach significantly improves the quality of OD datasets, particularly those used in state-of-the-art models. A visual representation of the proposed framework is illustrated in Figure~\ref{Framework}, and the procedures for each stage are outlined in Algorithm 1 and Algorithm 2. Each module of the framework is discussed in detail below.

\subsection{Data Preparation and Annotations Error Detection}
The MS-COCO \cite{lin2014microsoft} dataset, a widely recognized benchmark for OD research, contains various annotation errors, such as missing annotations, bounding box inconsistencies, incorrect labels, grouping errors, and other inaccuracies that can significantly impact model training and evaluation. As a preliminary step in developing our algorithm, referred to as data preparation, a comprehensive evaluation of Ground Truth (GT) annotations was conducted by a human expert, termed the Oracle. Instead of categorizing errors by type, Oracle focused on identifying all images containing inaccurately annotated data across the entire dataset. These inaccurately annotated images are defined as those exhibiting errors such as incorrect bounding boxes, mislabeled objects, missed objects, or groupings, and they are called one object. They are distinguished from accurately annotated images, which consist of images with precise and complete correct annotations. This human verification process established a high-quality GT reference, forming the foundational stage of the data preparation phase. This phase is essential for achieving multiple critical objectives: assessing dataset quality for each object in the image, validating the annotation error detection models, and analyzing the impact of annotation errors on model performance. This process is visualized in Figure 2 (Step 1), where a dataset called COCO T/F from the original MS-COCO. The data preparation process of identifying inaccurately annotated images lays a robust groundwork for enhancing the quality of the MS-COCO dataset and optimizing subsequent algorithm development. 
\begin{algorithm}
\caption{\textcolor{black}{Annotations Error Detection}}
\begin{algorithmic}[1]

\Require $\mathcal{D}_{\text{COCO}}=\{(x_i, l_i, b_i)\}_{i=1}^{n}$, $\mathcal{D}_{\text{TF}}=\{(x_j,y_j)\}_{j=1}^{m}$
\Ensure $\mathcal{D}_{\epsilon}$ (Detected Label Errors)

\State $\mathcal{D}_{\epsilon} \gets \emptyset$

\For{$(x, l, b) \in \mathcal{D}_{\text{COCO}}$}
    \State $\mathcal{R}_x \gets (\text{Faster-RCNN})(x)$
    \State $C_x \gets f_{\text{cls}}(\mathcal{R}_x)$
    \State $(\nabla,\mathcal{L}) \gets (\nabla_x(C_x),\mathcal{L}(C_x))$
    \State $s_x\gets f_{\text{unsup}}(\nabla,\mathcal{L})$
    \If{$s_x = \epsilon$}
        \State $\mathcal{D}_{\epsilon}\gets\mathcal{D}_{\epsilon}\cup\{(x,l,b)\}$
    \EndIf
\EndFor

\State \Return $\mathcal{D}_{\epsilon}$

\end{algorithmic}
\end{algorithm}
The proposed automatic label error detection process leverages the unified architecture of Faster R-CNN, integrates with an annotation error detection module into a single, end-to-end trainable model. This framework focuses on optimizing regression loss for accurate bounding box localization, enabling the effective identification of annotation errors by analyzing deviations from expected learning patterns. The region proposal generation begins with anchor initialization, where predefined anchor boxes are placed across feature maps with strides of $\{8, 16, 32, 64, 128\}$. These anchors are defined at multiple scales $\{32^2, 64^2, 128^2, 256^2, 512^2\}$ and aspect ratios $\{1{:}1, 2{:}1, 1{:}2\}$, allowing the model to effectively detect objects of various sizes and shapes. Each anchor is centered on a sliding window location, and proposals are generated by regressing from these reference anchors. Feature extraction is performed using a convolutional backbone (e.g., ResNet-50), with shared convolutional layers feeding both the RPN and the detection head. The RPN slides a small network over the convolutional feature map to predict both objectness scores and bounding box refinements. This proposal mechanism is fully convolutional and translation-invariant, making it computationally efficient and well-suited for dense region proposals. The region proposals are scored and ranked using Non-Maximum Suppression (NMS), and the top-ranked proposals are forwarded to the Faster R-CNN head for further refinement and classification. The loss function in Faster R-CNN is a multitask loss comprising classification and regression terms. The regression loss is computed for positive anchors and measures the bounding box alignment using a smooth $L_1$ loss, indirectly guided by the Intersection-over-Union (IoU) between predicted and ground-truth boxes \textcolor{black}{is defined in Eq. 1:}

\begin{equation}
    \text{IoU} = \frac{A \cap B}{A + B - A \cap B}
\end{equation}

where $A$ and $B$ denote the predicted and ground-truth bounding box areas, respectively. By minimizing this regression loss across training iterations, the bounding box predictions become increasingly accurate.
After generating proposals, regression losses are normalized to decouple them from objectness loss, ensuring that loss signals remain stable even when no object is present in the GT. This stabilization is essential for detecting annotation inconsistencies, as it ensures reliable monitoring of label deviations regardless of object presence.
\textcolor{black}{Therefore, this study introduces a modified regression loss based on a normalization mechanism to ensure stability and enhance the detection of erroneous annotations.}
In conventional OD neural network optimization typically involves minimizing a loss function through gradient-based backpropagation. Samples that consistently produce high loss values and large gradients often indicate learning difficulties, potentially resulting from annotation errors or inherently challenging data. This study proposes a systematic approach to detect such anomalous samples by tracking loss and gradient metrics in real-time throughout the training process.

In conventional OD, the regression loss is typically normalized only by the number of correctly matched ground-truth objects within an image. However, This approach neglects the information conveyed by false positives, which are predicted bounding boxes that do not match any ground-truth object. To address this, we introduce a refined normalization scheme for regression loss, \textcolor{black}{defined in Eq 2:}

\begin{equation}
L' = \frac{L}{N_{\text{matching}(gt, pred)}} \cdot (1 + N_{\text{FP}})
\end{equation} 
where \( L \) represents the original regression loss, \( N_{\text{matching}(gt, pred)} \) denotes the number of correctly matched bounding boxes, and \( N_{\text{FP}} \) is the number of false-positive bounding boxes predicted by the model. By incorporating false positives into the normalization factor, this formulation more accurately captures total prediction errors and enables better identification of difficult or anomalous samples. An additional challenge occurs when the model fails to predict any bounding boxes for images containing valid ground-truth annotations. Under conventional approaches, this scenario results in a regression loss of zero, potentially masking detection failures. To preserve meaningful loss information in such cases, we propose an exception-handling method \textcolor{black}{in which} zero-valued regression losses for images with valid ground-truth annotations are substituted with the maximum recorded loss during training, \textcolor{black}{as defined in Eq. 3:}
\begin{equation}
L_{\text{adjusted}} = 
\begin{cases}
\max(L_{\text{history}}), & \text{if } L = 0 \text{ and } N_{\text{GT}} > 0 \\[6pt]
L, & \text{otherwise}
\end{cases}
\end{equation}

This adjustment ensures that detection failures are accurately represented, facilitating improved identification of problematic samples.
To thoroughly assess learning difficulty, both loss values and gradient magnitudes are continuously monitored during training. Specifically, the gradient of the loss function with respect to the model parameters \(\theta\) at training step \(t\) \textcolor{black}{is defined in Eq. 4:}

\begin{equation}
g_t = \nabla_\theta L(x; \theta)
\end{equation}

where \( L(x; \theta) \) represents the loss computed for a given input sample \( x \). Since OD models often employ dynamic learning rates, gradient magnitudes can fluctuate substantially over time, complicating consistent interpretation. To address this, a normalization factor based on the learning rate schedule is implicitly applied during monitoring to ensure fair comparisons of gradient behavior across different training phases. This normalization ensures that gradient magnitudes remain comparable and consistent throughout training, facilitating accurate identification of anomalous samples. Real-time monitoring includes four distinct loss components and their corresponding gradient magnitudes for each image sample during training. The monitored losses are: the RPN classification loss (\( \text{loss\_rpn\_cls} \)), the RPN bounding box regression loss (\( \text{loss\_rpn\_bbox} \)), the ROI head classification loss (\( \text{loss\_cls} \)), and the ROI head bounding box regression loss (\( \text{loss\_bbox} \)). Correspondingly, gradient magnitudes associated with these loss components (\( \text{grad\_loss\_rpn\_cls} \), \( \text{grad\_loss\_rpn\_bbox} \), \( \text{grad\_loss\_cls} \), and \( \text{grad\_loss\_bbox} \)) are also recorded. Continuous tracking of these metrics enables dynamic evaluation of each sample's learning difficulty and contribution to model optimization, thus aiding in detecting persistent anomalies or annotation errors. Finally, given that unsupervised anomaly detection is impractical due to the absence of explicit anomaly labels in OD datasets, this study adopts an unsupervised approach utilizing a linear \textcolor{black}{AutoEncoder} model. The normalized loss and gradient values form an 8-dimensional input vector to the linear \textcolor{black}{AutoEncoder}, which learns to reconstruct these inputs during training. Annotations error as anomalous patterns are subsequently identified based on the reconstruction error, \textcolor{black}{defined in Eq. 5:}
\begin{equation}
\mathcal{L}_{\text{recon}} = \|\mathbf{z} - \hat{\mathbf{z}}\|^2
\end{equation}
Here, \(\mathbf{z}\) is the original 8-dimensional input feature vector consisting of loss and gradient metrics, and \(\hat{\mathbf{z}}\) is the \textcolor{black}{AutoEncoder's} reconstructed output. Samples exhibiting high reconstruction errors deviate significantly from typical training patterns, signaling potential anomalies. The adoption of a linear \textcolor{black}{AutoEncoder} thus enables effective detection of anomalous data without the requirement for explicit anomaly labels, enhancing the robustness and stability of the model training process. Moreover, the selection of a two-stage object detector such as Faster R-CNN, further facilitates annotation errors detection. Unlike one-stage models such as YOLO \cite{ge2021yolox} or RetinaNet \cite{lin2017focal}, two-stage frameworks clearly separate the region proposal and classification stages, allowing more granular and interpretable analysis of localization versus classification failures, thus improving overall anomaly identification accuracy.
To further enhance annotations errors detection reliability, two weighting strategies were proposed to balance the contributions of loss and gradient components. The first, termed global weighting, uniformly weights all eight monitored features. The second assigns individual weights to each component, allowing differential emphasis based on training dynamics. The optimal weighting was empirically determined after experiments shown in the results section to be $\lambda_{\text{loss}} = 0.9$ and $\lambda_{\text{grad}} = 0.1$, highlighting the greater stability and informativeness of loss values in representing anomalous behavior. This approach enables the anomaly detection model to be more sensitive to hard-to-learn samples without overreacting to transient gradient fluctuations.
\begin{algorithm}
\caption{\textcolor{black}{Pseudo Labeling}}
\begin{algorithmic}[1]

\Require $\mathcal{D}_{\epsilon}$ (Label Errors)
\Ensure $\mathcal{D}_{\text{reannotated}}$

\State $\mathcal{D}_{\text{reannotated}}\gets\emptyset$

\For{$(x_e,l_e,b_e)\in\mathcal{D}_{\epsilon}$}

    \State  $\mathcal{X}_{\text{aug}}\gets\{f_{\text{inv}}(x_e)\}_{k=1}^{6}$
    \State $\{(b_k,l_k,s_k)\}_{k=1}^{6}\gets f_{\text{det}}(\mathcal{X}_{\text{aug}})$
    \If{$|\{l_k:l_k=\tilde{l}\}|\geq4,\exists \tilde{l}$}
    
    \State $\tilde{l} \gets \text{argmax}_{l_k} \{s_k\}$
    \State $(b', l') \gets \left( \frac{1}{|\{l_k = \tilde{l}\}|} \sum_{l_k = \tilde{l}} b_k,\ \tilde{l} \right)$

    \Else
        \State $\textbf{continue}$
    \EndIf

    \State $\mathcal{B}_r\gets\{(b',l',s')|\text{IoU}(b'_a,b'_b)<0.6\lor\max(s'_a,s'_b)\}$

    \State $\mathcal{B}_v\gets\emptyset$
    \For{$(b_r,l_r,s_r)\in\mathcal{B}_r$}
    \State $(P_{3},P_{1},p_{c})\gets f_{\text{cls-img}}(x_e[b_r])$
    \If{$s_r\geq0.6\lor(0.3\leq s_r<0.6\land l_r\in P_{3})\lor$}
    \Statex $\quad(0.2\leq s_r<0.3\land l_r=P_{1})\lor$
    \Statex $\quad(0.1\leq s_r<0.2\land l_r=P_{1}\land p_c\geq0.8)\lor$
    \Statex $\quad(s_r<0.1\land l_r=P_{1}\land p_c\geq0.9)$
        \State $\mathcal{B}_v\gets\mathcal{B}_v\cup\{(b_r,l_r)\}$
    \EndIf
    \EndFor

    \State
    \For{$(b_v,l_v)\in\mathcal{B}_v$}
        \State $G\gets\text{Grad-CAM}(x_e[b_v],l_v)$
        \If{$\text{Var}(G)\leq\alpha$}
            \State $\mathcal{D}_{\text{reannotated}}\gets\mathcal{D}_{\text{reannotated}}\cup\{(x_e,b_v,l_v)\}$
        \ElsIf{$\text{Var}(G)>\alpha\land\text{conc}(G)\geq\beta$}
            \State $(b^*,l^*)\gets f_{\text{opt}}(b_v,G)$
            \State $\mathcal{D}_{\text{reannotated}}\gets\mathcal{D}_{\text{reannotated}}\cup\{(x_e,b^*,l^*)\}$
        \EndIf
    \EndFor
\EndFor

 $\mathcal{D}_{\text{reannotated}}\gets\mathcal{D}_{\text{reannotated}}\cup\{(x_t,\emptyset,\emptyset)\}$

\State \Return $\mathcal{D}_{\text{reannotated}}$

\end{algorithmic}
\end{algorithm}

\subsection{Pseudo Labeling and Evaluation }

The proposed pseudo-labeling process involves a systematic four-stage refinement approach aimed at generating highly reliable pseudo-labels from the erroneously annotated data identified during the annotation error detection phase.
Unlike traditional pseudo-labeling techniques that directly utilize bounding boxes predicted by OD models without further validation, this process employs a multi-step verification mechanism to enhance bounding box reliability. The entire process, as thoroughly detailed and outlined in Algorithm 2, ensures that errors introduced during the initial detection phase are progressively corrected through robust filtering and refinement techniques.
The initial stage involves generating multiple bounding boxes for a single object instance using a technique referred to as an invertible transformation. OD models typically produce a single prediction per input image; however, applying various augmentations can result in slightly different detections for the same object. To exploit this characteristic, six different transformations are applied to each input image: original, vertical flip, horizontal flip, upscaling with horizontal flip, upscaling with vertical flip, and downscaling. The objective is to create a diverse set of bounding boxes for a single object instance. The resulting bounding boxes, class labels, and confidence scores are generated by applying the detection model to each transformed image from the set of augmented images \textcolor{black}{defined in Eq. 6:}

\begin{equation}
    \mathcal{X}_{\text{aug}} = \{f_{\text{inv}}(x_e)\}_{k=1}^{6}
\end{equation}

The predictions corresponding to these transformed images are explicitly~\textcolor{black}{defined in Eq. 7:} 

\begin{equation}
    \{(b_k, l_k, s_k)\}_{k=1}^{6} = f_{\text{det}}(\mathcal{X}_{\text{aug}})
\end{equation}
Where \(b_k\) represents the bounding box, \(l_k\) the class label, and \(s_k\) the confidence score for each transformation \(k\). 

To ensure consistency across predictions, the label corresponding to the highest confidence score among the transformed images is selected. Instead of relying on majority voting, this approach identifies the most confident prediction and then averages all bounding boxes that share its label to generate a stable pseudo label. This strategy reduces the risk of including erroneous detections caused by individual transformations while enhancing robustness through spatial averaging, as~\textcolor{black}{is defined in Eq. 8:}

\begin{equation}
(b', l') = \left( \frac{1}{|\{l_k = \tilde{l}\}|} \sum_{l_k = \tilde{l}} b_k,\ \tilde{l} \right),\quad \tilde{l} = \text{argmax}_{l_k} \{s_k\}
\end{equation}

where \(\tilde{l}\) is the class label corresponding to the highest confidence score among all predictions. The second refinement stage focuses on eliminating redundant bounding boxes corresponding to the same object through IoU-based filtering. Instead of using NMS, which discards overlapping boxes purely based on confidence ranking, this approach applies a more nuanced method by comparing bounding boxes using the IoU, as defined in Eq. 1. Bounding boxes with an IoU of 0.6 or higher are considered to refer to the same object. Among these overlapping boxes, only the one with the highest confidence score is retained, while the rest are removed. This filtering process \textcolor{black}{is defined in Eq. 9:}

\begin{equation}
    \mathcal{B}_r = \{(b', l', s') \mid \text{IoU}(b'_a, b'_b) < 0.8 \lor \max(s'_a, s'_b)\}
\end{equation}

The use of a high IoU threshold of 0.8 ensures that only highly overlapping boxes are filtered, thereby retaining the most reliable predictions. The refined bounding boxes are saved separately for further validation.
The third refinement stage aims to further validate the bounding boxes by employing a classification model. OD models use confidence scores to indicate the reliability of detected objects, but these scores alone are not always sufficient, particularly for small objects or those situated in complex backgrounds. To address this limitation, a ResNet-50-based classification model is applied to the bounding boxes with lower confidence scores. The bounding boxes are cropped from the original images and fed into the classification model. The model produces a set of predictions consisting of the Top-3 predicted labels \( P_{3} \), the Top-1 predicted label \( P_{1} \), and the associated probability \( p_c \) as given in the Algorithm 2. The validation process is structured based on the confidence score of the bounding boxes. If a bounding box has a confidence score \( s_r \geq 0.5 \), it is retained without further validation, as it is considered sufficiently reliable. For bounding boxes with a confidence score between \( 0.3 \leq s_r < 0.5 \), the box is retained if the original class appears in the model’s Top-3 predictions, ensuring that the detected object is likely to be correctly identified even if the confidence score is moderately low. When the confidence score falls within the range \( 0.2 \leq s_r < 0.3 \), the box is retained if the Top-1 prediction matches the original class, reflecting a stricter criterion due to the lower initial confidence. For confidence scores between \( 0.1 \leq s_r < 0.2 \), the box is retained only if the Top-1 prediction matches the original class and the associated probability \( p_c \) is at least \( 0.5 \). As the confidence score decreases further, for bounding boxes within the range \( 0.08 \leq s_r < 0.1 \), the box is retained if the Top-1 prediction matches the original class and the confidence score \( s_r \) is at least \( 0.6 \). Finally, for the lowest confidence scores \( s_r < 0.08 \), the bounding box is only retained if the Top-1 prediction matches the original class and has a confidence score of at least \( 0.7 \). This hierarchical validation strategy ensures that low-confidence predictions are accurately assessed before being discarded or retained, thereby minimizing the risk of retaining erroneous pseudo labels while preserving true positives that may have low confidence scores due to various challenges such as occlusion, complex backgrounds, or small object sizes.

The final stage applies Gradient-weighted Class Activation Mapping (Grad-CAM) to further refine the retained bounding boxes by visualizing the activation regions. This technique generates heatmaps highlighting areas within each bounding box that contribute most to the model’s predictions. The generated activation map \( G \) \textcolor{black}{is defined in Eq. 10:}

\begin{equation}
    G = \text{Grad-CAM}(x_e[b_v], l_v)
\end{equation}

Bounding boxes are validated or adjusted based on the variance and concentration of the activation regions. The refinement ensures that the final pseudo labels are accurately aligned with the underlying objects. The entire pseudo-labeling process effectively mitigates errors in bounding box generation by combining consistency checks, redundancy filtering, confidence-based validation, and visual explanation techniques. By employing this structured approach, the proposed method ensures the generation of high-quality pseudo labels, enhancing the training dataset's reliability.

The effectiveness of the re-labeled dataset, called MJ-COCO, was verified in terms of performance as discussed in the results section using cross-validation across multiple OD architectures without any change in the architecture, including RetinaNet, YOLOv3, YOLOX, Faster R-CNN, etc. Significant improvements were observed on external datasets such as Sama COCO \cite{zimmermann2023benchmarking}, which feature cleaner annotations. This indicates that the benefits of re-labeling are more pronounced, thereby validating the robustness and generalization potential of the proposed correction pipeline.

\renewcommand{\tablename}{Table}
{\renewcommand{\thetable}{\arabic{table}}
\section{experimental results}
 \label{sec:experiments_results}
 \subsection {Experimental Setting}
 All experiments were conducted on a high-performance computing setup comprising an Intel(R) Xeon(R) Gold 6230 CPU, dual NVIDIA RTX 3090 Ti GPUs, and 64 GB RAM (32 GB × 2), running on the Ubuntu 18.04 operating system. The implementation was carried out using the PyTorch deep learning framework, supplemented with additional libraries to support training and evaluation processes.
The proposed model was trained using the Stochastic Gradient Descent (SGD) optimizer, which updates model weights in mini-batch increments to minimize the loss function. The learning rate was set to 0.05, with a weight decay of 0.001 to prevent overfitting. A batch size of 32 and a momentum value of 0.9 were used throughout the training. The model was trained for 100 epochs to ensure convergence and optimal performance.
For all experiments involving the Region Proposal Network (RPN), we adopted ResNet-50 as the backbone network. Additionally, the number of cascade stages was fixed at two for consistency across all experimental configurations.
  \subsection {Dataset }

\begin{figure*}[htbp]
    \centering
    \includegraphics[width=\linewidth]{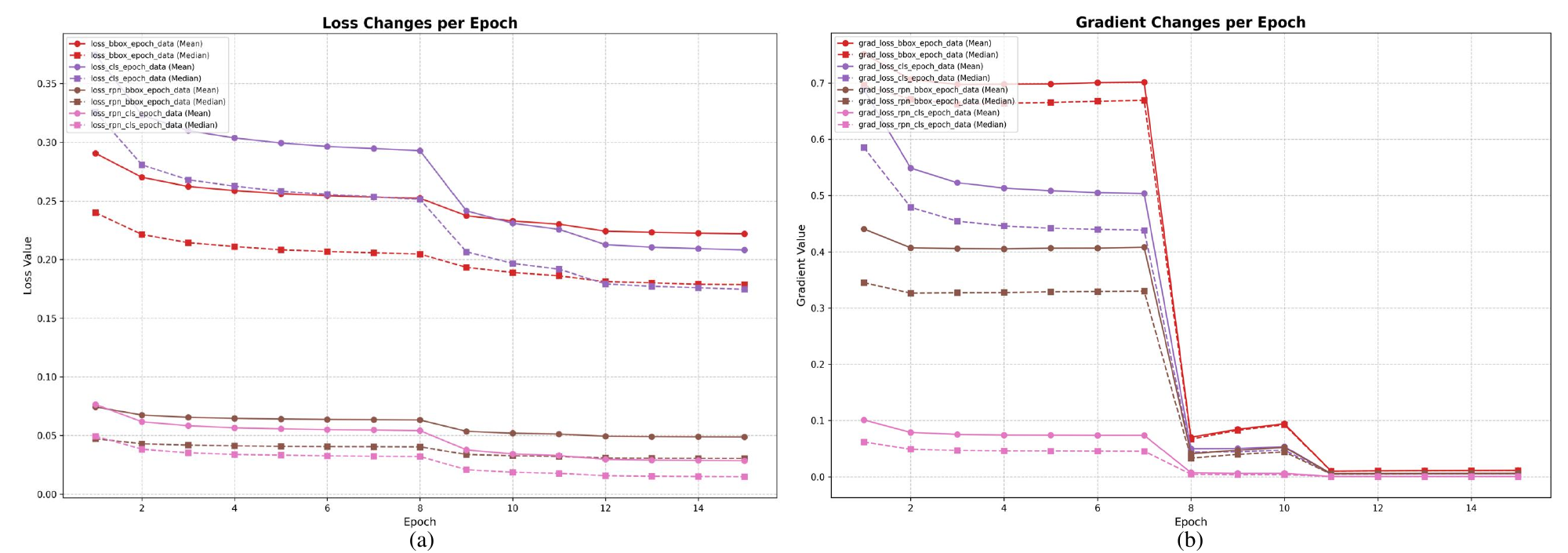}
    \caption{Visualization of loss and gradient variations across training epochs. (a) illustrates the changes in the mean and median of loss values per epoch, while (b) shows the corresponding changes in the mean and median of gradient values per epoch.}
    \label{Loss_GradientChange_Per_Epochs}
\end{figure*}
The MS-COCO dataset is a widely recognized benchmark for OD tasks. It contains over 330,000 images and more than 200,000 labeled instances across 80 object categories. Collected from complex real-world scenes, MS-COCO serves as a standard dataset for training and evaluating OD models. However, due to the inherent complexity of the scenes, it suffers from several annotation errors, as discussed in Section 1.
To identify these annotation errors across all 80 categories, we collected corrected annotations for the same classes from multiple existing datasets. To enhance diversity and maintain class balance, additional data were incorporated from Objects365 \cite{shao2019objects365}, PASCAL VOC \cite{everingham2011pascal}, Caltech256\cite{griffin2007caltech}, ImageNet\cite{deng2009imagenet}, CIFAR\cite{krizhevsky2010cifar}, and Roboflow\cite{ciaglia2022roboflow}. However, these datasets could not be directly integrated into the MS-COCO due to differences in class definitions. For instance, the Car class of MS-COCO is subdivided into Car, SUV, Sedan, and Sports Car in other Objects365. Similarly, in the MS-COCO the bird category is represented by bird while there are different names such as Wild Bird, Pigeon, and Penguin in Objects365. Likewise, the Sports Ball class in MS-COCO is also broken down into Tennis, Baseball, and Golf Ball in Objects365, with no direct equivalent in PASCAL VOC. For the Airplane class there are different names such as fighter-jet, airliner, military plane, etc. in Caltech256, ImageNet, and Roboflow. To address these inconsistencies, we conducted a class integration process based on the MS-COCO dataset taxonomy. Therefore, relevant data were extracted from these datasets only when the class definitions were identical and semantically aligned with those in MS-COCO.

\subsection {Evaluation Metrics}

To recognize annotation errors in the MS-COCO dataset for OD, we trained a classification model to detect mislabeled or incorrectly annotated instances which degrade the overall performance of the OD. The performance of the classification model is evaluated using standard metrics, including accuracy, precision, recall, and F1-score, which provide a comprehensive assessment of the model’s ability to distinguish between correctly and incorrectly annotated data. Their mathematical formulas \textcolor{black}{are defined in Eqs. 11 to 15.} Additionally, we evaluated performance using threshold-independent metrics, including the Area Under the Receiver Operating Characteristic Curve (AUROC) and the Area Under the Precision-Recall Curve (PRAUC). These metrics assess the model’s sensitivity and precision at various thresholds.

Accuracy represents the proportion of correctly classified instances among all predictions, providing an overall measure of model performance:%

\begin{equation}
\text{Accuracy} = \frac{TP + TN}{TP + TN + FP + FN}
\label{eq:accuracy}
\end{equation}

Precision indicates how many of the instances predicted as errors are actually errors, minimizing false positives:

\begin{equation}
\text{Precision} = \frac{TP}{TP + FP}
\label{eq:precision}
\end{equation}

Recall measures the model’s ability to detect all actual annotation errors, ensuring minimal false negatives:

\begin{equation}
\text{Recall} = \frac{TP}{TP + FN}
\label{eq:recall}
\end{equation}

Since precision and recall often trade off, the F1-Score balances them by computing their harmonic mean, offering a comprehensive evaluation of the model’s effectiveness in detecting annotation errors:

\begin{equation}
\text{F1-Score} = \frac{2 \times \text{Precision} \times \text{Recall}}{\text{Precision} + \text{Recall}}
\label{eq:f1-score}
\end{equation}

Where \textcolor{black}{True Positives (TP)} represents correctly detected annotation errors, \textcolor{black}{True Negatives (TN)} are correctly identified normal annotations, \textcolor{black}{False Positives (FP)} are normal annotations incorrectly classified as errors, and \textcolor{black}{False Negatives (FN)} are actual annotation errors that were not detected.
Furthermore, to assess OD model performance comprehensively, we utilized the Average Precision (AP) metric, computed across multiple IoU thresholds ranging from 0.50 to 0.95. AP provides an aggregate measure of precision at different IoU thresholds:
\begin{equation}
\text{AP} = \frac{1}{N} \sum_{i=1}^{N} \text{Precision}(\text{IoU}_i)
\label{eq:AP}
\end{equation}

\subsection {Performance of Annotation Error Detection}
To address annotation error detection, the proposed Faster-RCNN-based framework was effectively optimized for OD tasks. However, during training, some samples exhibited high loss values, which were often attributed to annotation errors. These included issues such as missing object labels, incorrect class assignments, inaccurate bounding boxes, duplicate labels, inconsistent labeling criteria, improper object estimations, group labeling, and labeler-induced inconsistencies as shown in Figure~\ref{COCO_Errors}. A particularly noticeable spike in loss values was observed when objects present in the image were absent from the GT Figure~\ref{Loss_GradientChange_Per_Epochs}, this results in significant variations in the loss and gradients per epoch, offering useful insights for detecting annotation inconsistencies. Therefore, an unsupervised anomaly detection method is developed by leveraging these learning indicators for annotation error detection. 
For comparative analysis, we conducted a comprehensive ablation study using different unsupervised anomaly detection models based on AutoEncoders and generative architectures. The goal was to analyze how well these models could detect annotation errors by leveraging the loss and gradient values extracted from the OD training process. To fairly evaluate the proposed method, we conducted experiments based on unnormalized (Loss and Gradients) and normalised (Loss and Gradients) as given in \textcolor{black}{Tables~\ref{tab:AnomalyDetection_UnnormalizedLoss_and_Gradients} and~\ref{tab:AnomalyDetection_NormalizedLoss_and_Gradients}}.

\begin{table}[t]
\setlength{\tabcolsep}{5pt}
\centering
\caption{Performance of Anomaly Detection Models using Unnormalized loss and Gradient.}

\label{tab:AnomalyDetection_UnnormalizedLoss_and_Gradients}
\begin{tabular}{lcccc}
\toprule
Baseline Model & Accuracy & Precision & Recall & F1-score \\
\midrule
Variational AutoEncoder      & 0.6695 & 0.5255 & 0.5280 & 0.5268 \\
CNN\_AutoEncoder             & 0.6295 & 0.4683 & 0.4706 & 0.4694 \\
LSTM\_AutoEncoder            & 0.6689 & 0.5245 & 0.5271 & 0.5258 \\
ConvLSTM\_AutoEncoder        & 0.6583 & 0.5094 & 0.5119 & 0.5106 \\
Transformer\_AutoEncoder     & 0.7384 & 0.6238 & 0.6268 & 0.6253 \\
AnoGAN                       & 0.3543 & 0.0751 & 0.0755 & 0.0753 \\
GANomaly                     & 0.6930 & 0.5591 & 0.5618 & 0.5604 \\
Diffusion                    & 0.7143 & 0.5895 & 0.5923 & 0.5909 \\
CNN\_Diffusion               & 0.7172 & 0.5936 & 0.5965 & 0.5950 \\
LSTM\_Diffusion              & 0.5863 & 0.4066 & 0.4086 & 0.4076 \\
ConvLSTM\_Diffusion          & 0.5848 & 0.4044 & 0.4064 & 0.4054 \\
Transformer\_Diffusion       & 0.7118 & 0.5858 & 0.5887 & 0.5872 \\
\textbf{AutoEncoder}         & \textbf{0.7490} & \textbf{0.6390} & \textbf{0.6421} & \textbf{0.6406} \\
\bottomrule
\end{tabular}
\end{table}

Table~\ref{tab:AnomalyDetection_UnnormalizedLoss_and_Gradients} represents the anomaly detection performance using unnormalized loss and gradient values, we observe a considerable variation in model effectiveness across different architectures. The AutoEncoder achieved the highest performance, with an accuracy of 0.7490 and an F1-score of 0.6406. This suggests that even without architectural complexity, a basic reconstruction-based model can effectively learn the underlying distribution of normal loss and gradient values, and detect deviations caused by annotation errors. The Transformer AutoEncoder achieved the second-best performance with an F1-score of 0.6253, indicating that the self-attention mechanism is particularly useful in capturing global contextual relationships in temporal sequences of gradients and loss. The Diffusion model also showed strong performance with an F1-score of 0.5909, outperforming GAN-based methods like GANomaly (0.5604) and AnoGAN (0.0753). Interestingly, AnoGAN performed the worst, likely due to its unstable training and lower sensitivity to subtle loss variations, which are crucial for detecting annotation anomalies. CNN, LSTM, and ConvLSTM-based AutoEncoders demonstrated moderate results, with F1-scores in the range of 0.4694 to 0.5258. These models struggle due to their limitations in modeling long-term dependencies or complex spatial-temporal correlations present in the gradient/loss sequences.
\begin{table}[b]
\setlength{\tabcolsep}{5pt}
\centering
\caption{Performance of Anomaly Detection Models using Normalized loss and Gradient.}
\label{tab:AnomalyDetection_NormalizedLoss_and_Gradients}
\resizebox{0.48\textwidth}{!}{
\begin{tabular}{lcccc}
\toprule
Baseline Model & Accuracy & Precision & Recall & F1-score \\
\midrule
Variational AutoEncoder      & 0.6618 & 0.5145 & 0.5145 & 0.5145 \\
CNN\_AutoEncoder             & 0.6558 & 0.5059 & 0.5083 & 0.5071 \\
LSTM\_AutoEncoder            & 0.6559 & 0.5061 & 0.5085 & 0.5073 \\
ConvLSTM\_AutoEncoder        & 0.6464 & 0.4924 & 0.4948 & 0.4936 \\
Transformer\_AutoEncoder     & 0.7459 & 0.6345 & 0.6376 & 0.6361 \\
AnoGAN                       & 0.3522 & 0.0721 & 0.0725 & 0.0723 \\
GANomaly                     & 0.6891 & 0.5535 & 0.5562 & 0.5549 \\
Diffusion                    & 0.6800 & 0.5404 & 0.5430 & 0.5417 \\
CNN\_Diffusion               & 0.7158 & 0.5916 & 0.5944 & 0.5930 \\
LSTM\_Diffusion              & 0.5736 & 0.3884 & 0.3903 & 0.3894 \\
ConvLSTM\_Diffusion          & 0.5749 & 0.3902 & 0.3921 & 0.3912 \\
Transformer\_Diffusion       & 0.7096 & 0.5827 & 0.5856 & 0.5841 \\
\textbf{AutoEncoder}                  & \textbf{0.7916} & \textbf{0.6999} & \textbf{0.7033} & \textbf{0.7016} \\
\bottomrule
\end{tabular}}
\end{table}
Table~\ref{tab:AnomalyDetection_NormalizedLoss_and_Gradients}, which utilizes normalized loss and gradient values, shows consistent improvement across nearly all models. Normalization enhances model learning by reducing variance and aligning data distributions, thereby improving anomaly detection sensitivity. The AutoEncoder again outperformed all models, this time achieving an accuracy of 0.7916 and an F1-score of 0.7016, showing an improvement of approximately 4.26\% in accuracy and 6.1\% in F1-score over the unnormalized setting. This substantial gain highlights that normalization helps the model better distinguish between typical and anomalous samples, especially in the presence of noisy gradients caused by annotation inconsistencies. The Transformer AutoEncoder maintained its second-best position with an F1-score of 0.6361, again validating its strong generalization capabilities in handling structured input sequences. The CNN diffusion model closely followed, achieving an F1-score of 0.5930 and demonstrating robustness in scenarios where spatial features interact with noise. Notably, models like ConvLSTM Diffusion, LSTM Diffusion, and AnoGAN remained the lowest performers, with F1-scores below 0.40 and accuracy values near or below 0.58. This is attributed to their inherent limitations: GAN-based methods tend to be less stable and harder to train, while LSTM-based Diffusion models do not effectively capture subtle local patterns in normalized sequences due to their sequential bias.

When comparing Table 1 and Table 2, it is clear that normalization plays a critical role in improving model performance across the board. The AutoEncoder model’s performance gain from an F1-score of 0.6406 to 0.7016 represents a relative improvement of ~6.1\%, while its accuracy improved by ~4.26\%. Likewise, the Transformer AutoEncoder showed an increase in F1-score from 0.6253 to 0.6361, reflecting the consistent benefit of normalized inputs. Overall, these findings confirm that the normalized loss and gradients allow anomaly detection models to generalize better and offer improved sensitivity in detecting annotation errors, with the AutoEncoder and Transformer-based models emerging as the most reliable across both settings.

Since the AutoEncoder achieved the highest performance in accurately identifying annotation errors, further evaluations were conducted to examine its robustness across varying thresholds. Table~\ref{tab:anomaly_detection_threshold_variable} summarizes the AUROC and PRAUC metrics for both the existing (unnormalized) method and the proposed (normalized) loss and gradient calculation method. The proposed normalization approach clearly enhances performance, improving AUROC from 0.8003 to 0.8436 and PRAUC from 0.6243 to 0.7114. This indicates that normalization substantially boosts the model’s sensitivity and consistency, particularly beneficial when dealing with imbalanced annotation errors. A visual comparison clearly demonstrates that normalized inputs improve AUROC and PRAUC values, as shown in Figure~\ref{ErrorDetectionEncoder_AUROC_PRAUC}.

\begin{table}[b]
\setlength{\tabcolsep}{12pt}
\centering
\caption{Anomaly Detection Performance using Loss and Gradients (Threshold-Variable Evaluation).}
\resizebox{0.49\textwidth}{!}{%
\begin{tabular}{lcccc}
\hline
Model & \multicolumn{2}{c}{Existing Method} & \multicolumn{2}{c}{Proposed Method} \\ 
\cline{2-5}
 & AUROC & PRAUC & AUROC & PRAUC \\ 
\hline
AutoEncoder & 0.8003 & 0.6243 & \textbf{0.8436} & \textbf{0.7114} \\ 
\hline
\end{tabular}
}
\label{tab:anomaly_detection_threshold_variable}
\end{table}

\begin{table}[ht]
\setlength{\tabcolsep}{12pt}
\centering
\caption{Performance Evaluation According to Loss and Gradient Weight Variation (AutoEncoder).}
\resizebox{0.48\textwidth}{!}{%
\label{tab:lambda_weight_variationLOSS}
\begin{tabular}{ccccc}
\hline
$\lambda$ & Accuracy & Precision & Recall & F1-score \\
\hline
0.1 & 0.7411 & 0.6278 & 0.6308 & 0.6293 \\
0.2 & 0.7673 & 0.6652 & 0.6684 & 0.6668 \\
0.3 & 0.7800 & 0.6833 & 0.6866 & 0.6849 \\
0.4 & 0.7874 & 0.6939 & 0.6973 & 0.6956 \\
0.5 & 0.7916 & 0.6999 & 0.7033 & 0.7016 \\
0.6 & 0.7943 & 0.7038 & 0.7072 & 0.7055 \\
0.7 & 0.7962 & 0.7065 & 0.7099 & 0.7082 \\
\hline
\end{tabular}}
\end{table}

\begin{figure*}
    \centering
    \includegraphics[width=1\linewidth]{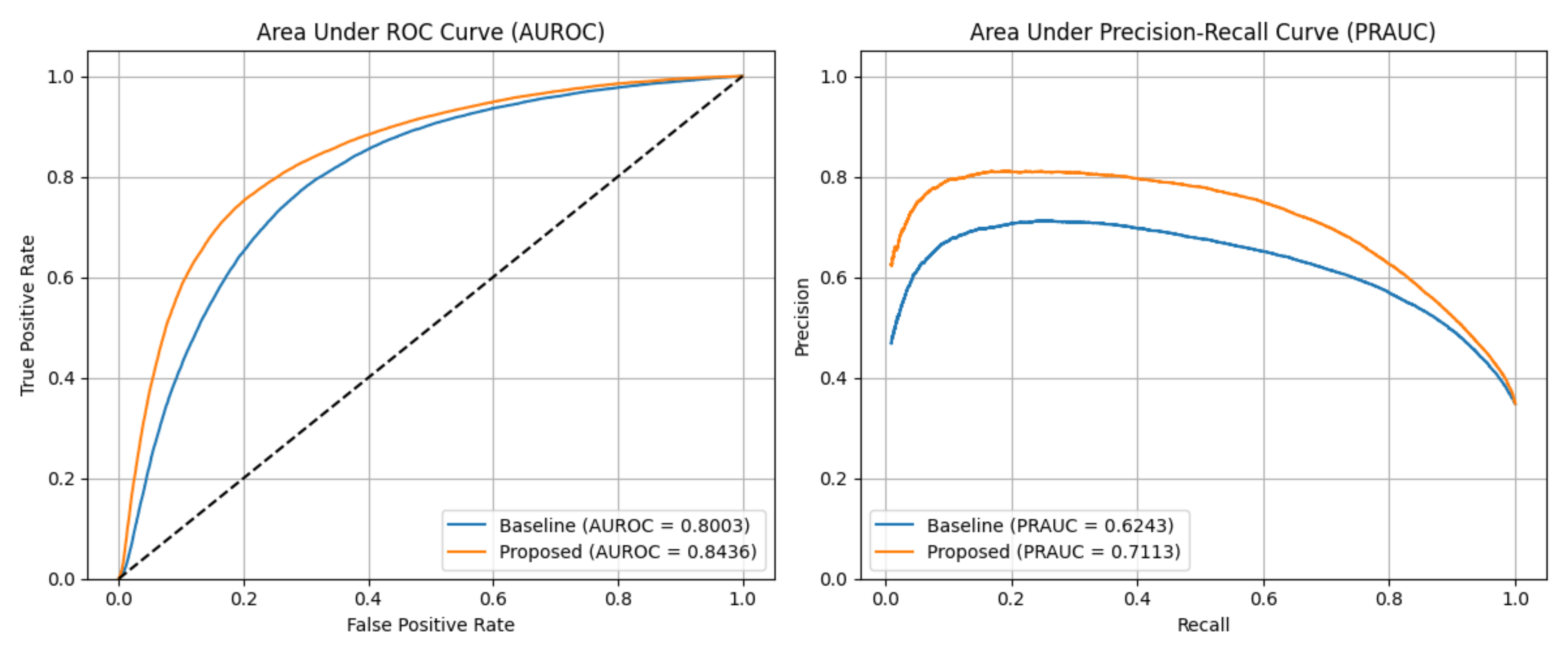}
    \caption{ Comparison of AUROC and PRAUC curves based on varying thresholds. The blue curve indicates the results of anomaly detection performed using unnormalized loss, while the yellow curve represents the results obtained by applying the proposed loss normalization method.}
    \label{ErrorDetectionEncoder_AUROC_PRAUC}
\end{figure*}

\begin{figure*}
    \centering
    \includegraphics[width=1\linewidth]{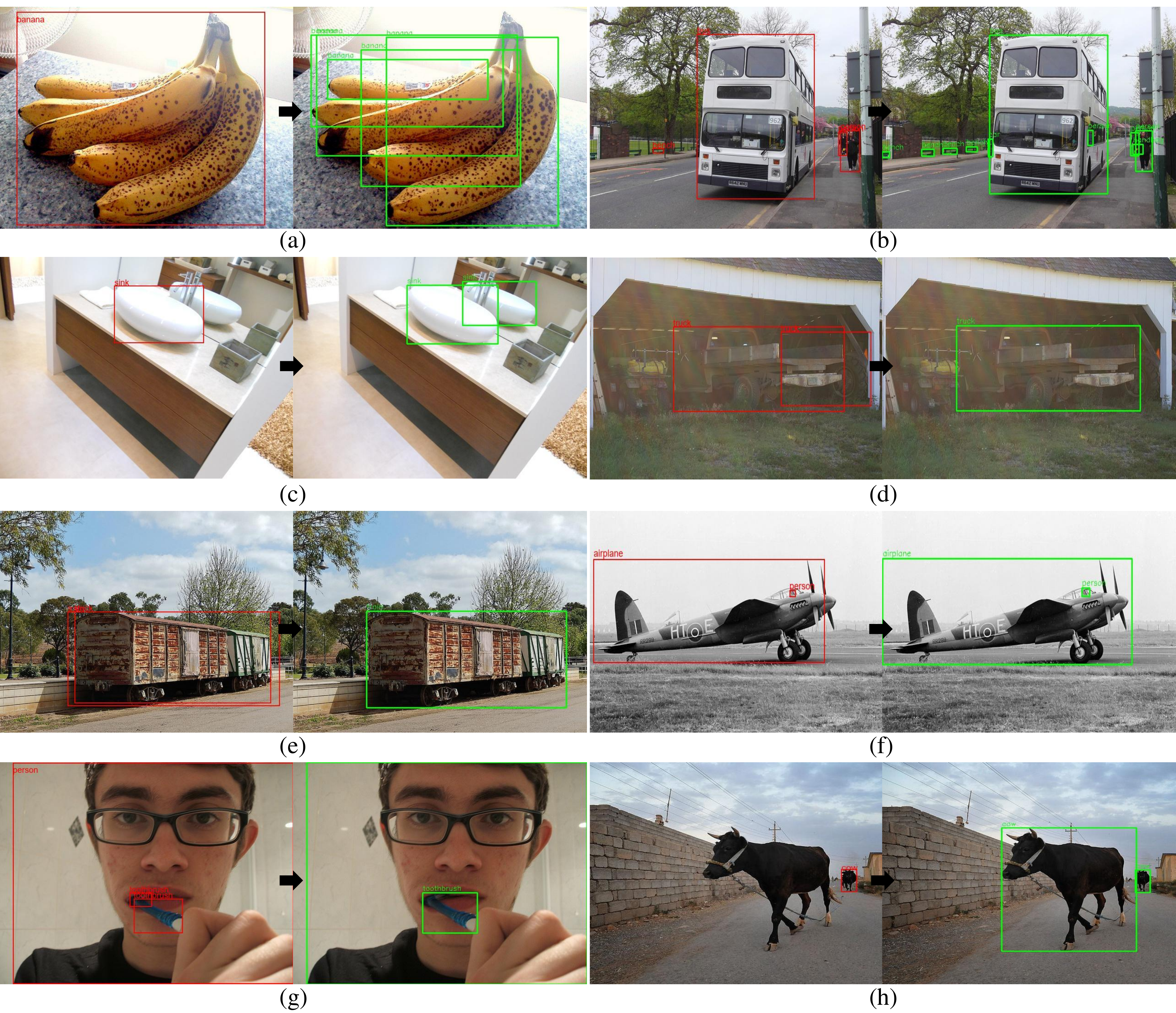}
    \caption{ Comparison of annotation errors in the MS-COCO dataset and corrections made by the proposed pseudo-labeling method. This figure presents side-by-side comparisons between object annotations from the original MS-COCO dataset (highlighted in red) and the corrected annotations generated by the proposed pseudo-labelling method (highlighted in green). (a) Group annotations correction, (b) Missing object labeling, (c) Incorrect object estimation, (d) Inaccurate annotation removal, (e) Duplicate annotation removal, (f) Bounding box optimization, (g) Bounding box correction, (h) Missing object annotation.}
    \label{PsudoLabellingResults}
\end{figure*}

Motivated by these improvements, we explored the impact of varying relative weighting the loss-to-gradient weighting ($\lambda$) to determine the most effective indicator for annotation errors. Here, $\lambda$ represents the emphasis on loss, while the gradient weight is $(1 - \lambda)$. Performance evaluations at a fixed threshold (top 35\%) were selected during the experiments, and their results are shown in 
 Table~\ref{tab:lambda_weight_variationLOSS}. Results indicated a consistent improvement as $\lambda$ increased, highlighting loss as a more reliable indicator for identifying annotation errors than gradients. The experimental outcomes indicate a clear trend wherein increasing the weight assigned to loss ($\lambda$) consistently improves anomaly detection performance across all metrics. Notably, the highest accuracy (0.7962) and F1-score (0.7082) were achieved at $\lambda = 0.7$. Based on these experiments that loss values serve as a more reliable and robust indicator of annotation inconsistencies than gradient values. \textcolor{black}{This was further verified by threshold-variable metrics such as AUROC and PRAUC, as given in supplementary file Table 1.} 
The AUROC and PRAUC evaluations corroborate the earlier findings from fixed-threshold analyses. Both metrics showed incremental improvements as $\lambda$ increased, with the highest AUROC (0.8534) and PRAUC (0.7378) observed at $\lambda = 0.9$. This outcome confirms that assigning a higher weight to loss enhances the model’s sensitivity and consistency in identifying annotation errors across varying anomaly thresholds.

Therefore, we conducted experiments to examined the influence of assigning individual weights ($\alpha$, $\beta$, $\gamma$, $\delta$) to specific loss components: RPN classification ($\alpha$), RPN bounding-box regression ($\beta$), Region of Interest (ROI) classification ($\gamma$), and ROI bounding-box regression ($\delta$). A comprehensive ablation study was performed to evaluate how combinations of these component-specific weights influence annotation error detection performance. \textcolor{black}{The experimental results, including the identification of the optimal combination of feature-specific weights at $\alpha = 0.3$, $\beta = 0.1$, $\gamma = 0.3$, and $\delta = 0.3$, yielded the highest overall performance metrics (accuracy: 0.7939, F1-score: 0.7048, AUROC: 0.8459, PRAUC: 0.7190), as given in supplementary file Table 2.}
This implies that annotation error detection performance benefits from a balanced yet differentiated emphasis on classification and regression tasks, particularly within the RPN and ROI stages. To fully leverage these findings, we conducted additional comparative experiments examining the performance improvements achievable through independent and combined weighting strategies (optimal $\lambda$ and $\alpha$–$\delta$ adjustments).
\textcolor{black}{Finally, to fully leverage these insights, we compared independent and combined weighting approaches (optimal $\lambda$ and feature-specific $\alpha$–$\delta$ adjustments), as given in supplementary file Table 3.}
The combined weighting strategy, integrating optimal loss-gradient weighting ($\lambda = 0.7$) with optimal feature-specific weights ($\alpha=0.3$, $\beta=0.1$, $\gamma=0.3$, $\delta=0.3$), consistently demonstrated superior anomaly detection performance, achieving the highest accuracy (0.8008), F1-score (0.7148), AUROC (0.8572), and PRAUC (0.7523). These findings strongly validate that comprehensive weight adjustment strategies are crucial in enhancing the sensitivity, precision, and overall stability of annotation error detection model.

Our findings indicate that out of the 117,267 analyzed images in the MS-COCO dataset, 64,645 images (55.1\%) were classified as normal \textcolor{black}{(TN)}, while 52,621 images (44.9\%) were identified as erroneous. Among these anomalous images, 29,267 were TP (correctly detected annotation errors), 11,776 were FP (normal images incorrectly flagged as anomalies), and 11,578 were FN (actual annotation errors missed by the model). These results highlight the significant presence of annotation errors in benchmark datasets and emphasize the necessity for robust annotation verification methods and relabeling of the dataset to enhance the reliability of OD models.
\subsection {Pseudo Labeling Results}
The proposed pseudo-labeling method was designed to systematically identify and correct annotation errors within the MS-COCO dataset, significantly enhancing annotation accuracy and reliability through a structured four-stage refinement pipeline. Unlike traditional pseudo-labeling methods that rely directly on raw predictions from OD models, our method integrates rigorous validation at multiple stages, resulting in substantial improvements in dataset quality.
In the first stage, multiple augmented versions of each original image were generated using invertible transformations, including horizontal flips, vertical flips, upscaling, and combinations thereof. Each transformed image was individually processed by the OD model, producing multiple candidate bounding boxes for the same object. Bounding boxes consistently detected across at least four of the six augmented views were selected as reliable initial candidates. This consistency-based selection process effectively mitigated errors related to object omissions and inconsistent labeling. Subsequently, in the second stage, these initial bounding boxes underwent further refinement through an IoU based merging procedure. Bounding boxes identified as significantly overlapping (\text{IoU}~$\geq$~0.8) were merged into single, consolidated annotations. Coordinates of merged bounding boxes were conservatively determined by taking the minimal and maximal extents from overlapping boxes, ensuring comprehensive object coverage. Confidence scores were averaged conservatively, thereby enhancing annotation reliability. This merging process successfully addressed errors arising from duplicate labeling, fragmentation, and poor localization, resulting in more precise annotations.

\begin{figure*}[t]
    \centering
    \includegraphics[width=1\linewidth]{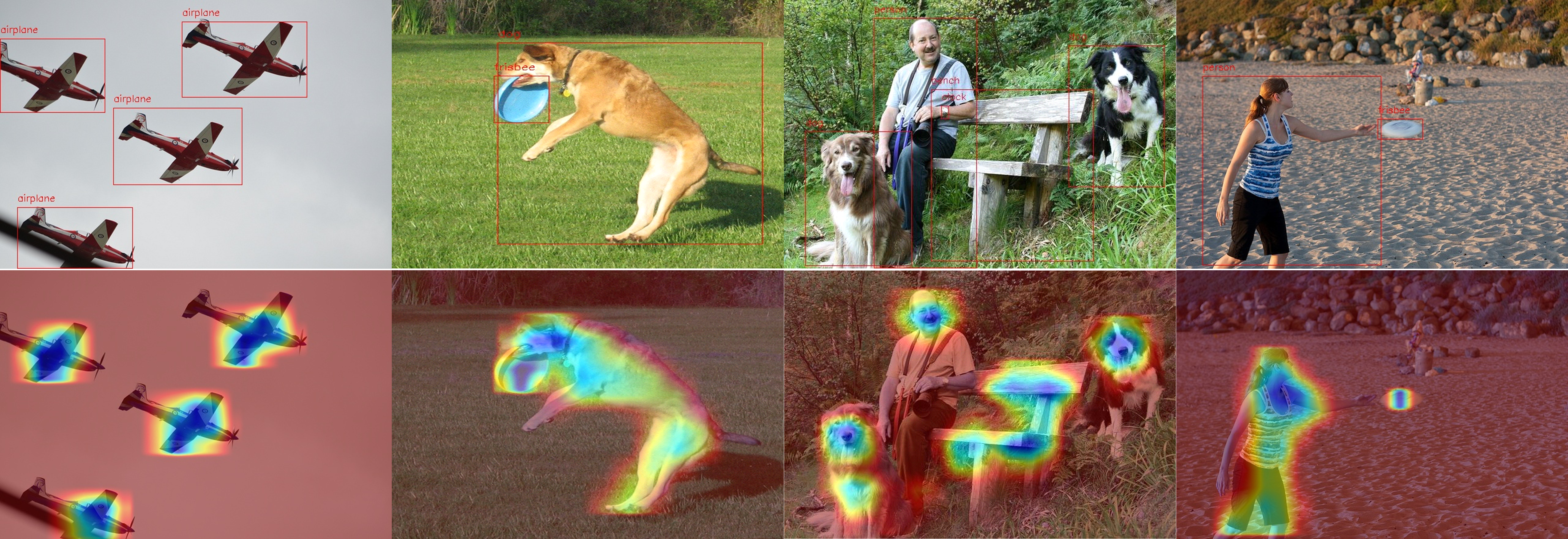}
    \caption{Bounding box refinement results based on activation regions using Grad-CAM. }
    \label{PsudoLabelling_CAM}
\end{figure*}

\begin{table}
\setlength{\tabcolsep}{5pt}
\centering
\caption{Comparison of Small Objects of the Original MS-COCO and the  Proposed MJ-COCO Dataset.}
\label{small_class_differences_with_area}
\begin{tabular}{lcccc}
\toprule
      Category &  MS-COCO &  MJ-COCO &  Difference &  Average Area \\
\midrule
Apple &     5851 &       19527 &       13676 &      11088.14 \\
Backpack &   8720 &       10029 &        1309 &       6856.27 \\
Baseball bat &   3276 &      3517 &         241 &       7594.75 \\
Baseball glove &     3747 &                  3440 &        -307 &       2764.16 \\
Book &    24715 &     35712 &       10997 &       5932.80 \\
Bottle &    24342 &     32455 &        8113 &       4451.28 \\
Car &    43867 &        51662 &        7795 &       7369.33 \\
Carrot &     7852 &        15411 &        7559 &       8332.94 \\
Cell phone &     6434 &     6642 &         208 &       8519.88 \\
Clock &     6334 &     7618 &        1284 &       8776.18 \\
Cup &    20650 &        22545 &        1895 &       6436.76 \\
Fork &     5479 &     5184 &        -295 &       9414.06 \\
Frisbee &     2682 &      2658 &         -24 &       5698.78 \\
Handbag &    12354 &      14524 &        2170 &       5749.00 \\
Kite &     9076 &      15092 &        6016 &       9455.52 \\
Knife &     7770 &      6697 &       -1073 &       6997.18 \\
Mouse &     2262 &        2377 &         115 &       4704.81 \\
Remote &     5703 &        5428 &        -275 &       5914.19 \\
Skateboard &     5543 &       5761 &         218 &       9310.90 \\
Skis &     6646 &         8945 &        2299 &       7569.37 \\
Spoon &     6165 &         6156 &          -9 &       6303.15 \\
Sports ball &     6347 &        6060 &        -287 &       1657.45 \\
Tennis racket &     4812 &     4932 &         120 &       9383.95 \\
Tie &     6496 &       6048 &        -448 &       9035.87 \\
Toothbrush &     1954 &      1901 &         -53 &       9172.48 \\
Traffic light &    12884 &      19583 &        6699 &       2342.25 \\
Wine glass &     7913 &      8429 &         516 &       7850.76 \\
\bottomrule
\end{tabular}
\end{table}

\begin{table}[!b]
\setlength{\tabcolsep}{5pt}
\centering
\caption{Comparison of Medium Scale Object between Original MS-COCO and Proposed Dataset.}
\label{medium_class_differences_with_area}
\begin{tabular}{lcccc}
\toprule
     Category &  MS-COCO &  MJ-COCO &  Difference &  Average Area \\
\midrule
       Banana &     9458 &                 49705 &       40247 &      18011.33 \\
        Bench &     9838 &                  9784 &         -54 &      23271.99 \\
      Bicycle &     7113 &                  7853 &         740 &      16589.63 \\
         Bird &    10806 &                 13346 &        2540 &      13040.90 \\
         Boat &    10759 &                 13386 &        2627 &      15401.33 \\
         Bowl &    14358 &                 13591 &        -767 &      23176.78 \\
     Broccoli &     7308 &                 14275 &        6967 &      16460.15 \\
        Chair &    38491 &                 56750 &       18259 &      11851.36 \\
          Cow &     8147 &                  8990 &         843 &      19902.50 \\
        Donut &     7179 &                 11622 &        4443 &      14948.45 \\
   Hair drier &      198 &                   202 &           4 &      12060.12 \\
     Keyboard &     2855 &                  3128 &         273 &      24034.52 \\
    Microwave &     1673 &                  1755 &          82 &      22134.03 \\
       Orange &     6399 &                 18416 &       12017 &      12336.81 \\
Parking meter &     1285 &                  1355 &          70 &      22190.20 \\
       Person &   262465 &                435252 &      172787 &      22284.61 \\
 Potted plant &     8652 &                 11252 &        2600 &      17280.37 \\
        Sheep &     9509 &                 12813 &        3304 &      13856.09 \\
         Sink &     5610 &                  5969 &         359 &      16088.43 \\
    Snowboard &     2685 &                  2565 &        -120 &      11093.93 \\
    Stop sign &     1983 &                  2684 &         701 &      22780.95 \\
    Surfboard &     6126 &                  6175 &          49 &      13284.59 \\
     Toaster &      225 &                   320 &          95 &      11882.80 \\
           TV &     5805 &                  6591 &         786 &      23429.79 \\
     Umbrella &    11431 &                 16895 &        5464 &      17625.26 \\
         Vase &     6613 &                  9684 &        3071 &      12706.60 \\
\bottomrule
\end{tabular}
\end{table}

\begin{table}
\setlength{\tabcolsep}{5pt}
\centering
\caption{ Comparison of Large Class Differences between Original MS-COCO and Proposed Dataset.}
\label{large_class_differences_with_area}
\begin{tabular}{lcccc}

\toprule
    \textbf{Category} &  MS-COCO &  MJ-COCO&  Difference &  Average Area \\
\midrule
    Airplane &     5135 &                  5810 &         675 &      51855.88 \\
        Bear &     1294 &                  1311 &          17 &      70436.14 \\
         Bed &     4192 &                  4177 &         -15 &     122376.85 \\
         Bus &     6069 &                  7132 &        1063 &      52573.34 \\
        Cake &     6353 &                  8968 &        2615 &      30014.28 \\
         Cat &     4768 &                  4895 &         127 &      73613.15 \\
       Couch &     5779 &                  5598 &        -181 &      62755.75 \\
Dining table &    15714 &                 16569 &         855 &     102777.17 \\
         Dog &     5508 &                  5870 &         362 &      50019.64 \\
    Elephant &     5513 &                  6233 &         720 &      46862.39 \\
Fire hydrant &     1865 &                  1877 &          12 &      31667.11 \\
     Giraffe &     5131 &                  5467 &         336 &      57309.55 \\
       Horse &     6587 &                  7120 &         533 &      36327.40 \\
     Hot dog &     2918 &                  3323 &         405 &      27108.37 \\
      Laptop &     4970 &                  5280 &         310 &      39638.33 \\
  Motorcycle &     8725 &                 10045 &        1320 &      35192.12 \\
        Oven &     3334 &                  4310 &         976 &      47937.49 \\
       Pizza &     5821 &                  6049 &         228 &      63539.53 \\
Refrigerator &     2637 &                  2728 &          91 &      58027.00 \\
    Sandwich &     4373 &                  3925 &        -448 &      41521.44 \\
    Scissors &     1481 &                  1558 &          77 &      25488.86 \\
    Suitcase &     6192 &                  7447 &        1255 &      26255.47 \\
  Teddy bear &     4793 &                  6432 &        1639 &      37882.29 \\
      Toilet &     4157 &                  4433 &         276 &      38209.85 \\
       Train &     4571 &                  4883 &         312 &      79644.79 \\
       Truck &     9973 &                 11476 &        1503 &      30036.18 \\
       Zebra &     5303 &                  6363 &        1060 &      39846.99 \\
\bottomrule
\end{tabular}
\end{table}

\begin{table*}
\setlength{\tabcolsep}{14pt}
\centering
\caption{Cross Validation of MJ-COCO using RetinaNet\cite{lin2017focal}.}
\label{tab:retinanet_comparison}
\resizebox{1.0\textwidth}{!}{%
\begin{tabular}{llcccccc}
\hline
Testing Data & Training Data & AP & AP$_{50}$ & AP$_{75}$ & AP$_S$ & AP$_M$ & AP$_L$ \\
\hline
\multirow{2}{*}{MS-COCO Validation \cite{lin2014microsoft}} 
& MS-COCO & \textbf{0.367} & \textbf{0.557} & \textbf{0.392} & \textbf{0.205} & \textbf{0.405} & \textbf{0.473} \\
& MJ-COCO  & 0.349 & 0.532 & 0.370 & 0.202 & 0.381 & 0.441 \\
\hline
\multirow{2}{*}{Sama Validation \cite{zimmermann2023benchmarking}} 
& MS-COCO & 0.356 & 0.532 & 0.379 & 0.198 & \textbf{0.401} & \textbf{0.484} \\
& MJ-COCO  & \textbf{0.358} & \textbf{0.538} & \textbf{0.381} & \textbf{0.208} & 0.399 & 0.466 \\
\hline
\multirow{2}{*}{Objects365 Validation \cite{shao2019objects365}} 
& MS-COCO & 0.221 & 0.336 & 0.235 & 0.078 & 0.207 & 0.335 \\
& MJ-COCO    & \textbf{0.227} & \textbf{0.345} & \textbf{0.243} & \textbf{0.081} & \textbf{0.217} & \textbf{0.339} \\
\hline
\multirow{2}{*}{PASCAL Validation\cite{everingham2011pascal} } 
& MS-COCO & \textbf{0.535} & \textbf{0.784} & \textbf{0.581} & \textbf{0.201} & \textbf{0.428} & \textbf{0.613} \\
& MJ-COCO    & 0.523 & 0.774 & 0.567 & 0.191 & 0.425 & 0.601 \\
\hline
\end{tabular}%
}
\end{table*}

\begin{table*}
\setlength{\tabcolsep}{14pt}
\centering
\caption{Cross Validation of MJ-COCO using YOLOv3\cite{redmon2018yolov3}.}
\label{tab:yolov3_comparison}
\resizebox{1.0\textwidth}{!}{%
\begin{tabular}{llcccccc}
\hline
Testing Data & Training Data & AP & AP$_{50}$ & AP$_{75}$ & AP$_S$ & AP$_M$ & AP$_L$ \\
\hline
\multirow{2}{*}{MS-COCO Validation \cite{lin2014microsoft}} 
& MS-COCO  & \textbf{0.294} & \textbf{0.513} & \textbf{0.299} & \textbf{0.114} &\textbf{ 0.325} & \textbf{0.452} \\
& MJ-COCO   & 0.286 & 0.500 & 0.286 & 0.110 & 0.317 & 0.431 \\
\hline
\multirow{2}{*}{Sama Validation \cite{zimmermann2023benchmarking}} 
& MS-COCO & 0.281 & 0.484 & 0.287 & 0.104 & 0.315 & \textbf{0.458} \\
& MJ-COCO   & \textbf{0.286} & \textbf{0.499} & \textbf{0.290} & \textbf{0.109} & \textbf{0.319} & 0.455 \\
\hline
\multirow{2}{*}{Objects365 Validation \cite{shao2019objects365}} 
& MS-COCO & 0.170 & 0.301 & 0.174 & 0.033 & 0.146 & 0.297 \\
& MJ-COCO    & \textbf{0.177} & \textbf{0.314} & \textbf{0.179} & \textbf{0.036} & \textbf{0.152} & \textbf{0.305} \\
\hline
\multirow{2}{*}{PASCAL Validation\cite{everingham2011pascal} } 
& MS-COCO  & \textbf{0.473} & 0.748 & 0.511 & 0.111 & \textbf{0.350} & \textbf{0.571} \\
& MJ-COCO   & 0.472 & \textbf{0.751} & \textbf{0.514} & \textbf{0.126} & 0.346 & 0.570 \\
\hline
\end{tabular}%
}
\end{table*}

\begin{table*}
\setlength{\tabcolsep}{14pt}
\centering
\caption{Cross Validation of MJ-COCO using YOLOX\cite{ge2021yolox}.}
\label{tab:yolox_comparison}
\resizebox{1.0\textwidth}{!}{%
\begin{tabular}{llcccccc}
\hline
Testing Data & Training Data & AP & AP$_{50}$ & AP$_{75}$ & AP$_S$ & AP$_M$ & AP$_L$ \\
\hline
\multirow{2}{*}{MS-COCO Validation \cite{lin2014microsoft}} 
& MS-COCO & \textbf{0.453} & \textbf{0.639} & \textbf{0.491} & \textbf{0.280} & \textbf{0.497} & \textbf{0.594} \\
& MJ-COCO    & 0.433 & 0.610 & 0.472 & 0.257 & 0.479 & 0.568 \\
\hline
\multirow{2}{*}{Sama Validation \cite{zimmermann2023benchmarking}} 
& MS-COCO  & 0.447 & 0.618 & 0.483 & 0.271 & 0.496 & 0.614 \\
& MJ-COCO    & \textbf{0.457} & \textbf{0.632} & \textbf{0.493} & \textbf{0.282} & \textbf{0.509} & \textbf{0.607} \\
\hline
\multirow{2}{*}{Objects365 Validation \cite{shao2019objects365}} 
& MS-COCO  & 0.288 & 0.402 & 0.311 & 0.110 & 0.272 & 0.428 \\
& MJ-COCO   & \textbf{0.302} & \textbf{0.417} & \textbf{0.326} & \textbf{0.119} & \textbf{0.287} & \textbf{0.439} \\
\hline
\multirow{2}{*}{PASCAL Validation\cite{everingham2011pascal} } 
& MS-COCO  & 0.624 & 0.832 & 0.679 & \textbf{0.237} & \textbf{0.504} & 0.708 \\
& MJ-COCO   & \textbf{0.629} & \textbf{0.834} & \textbf{0.691} & 0.232 & 0.503 & \textbf{0.714} \\
\hline
\end{tabular}%
}
\end{table*}

\begin{table*}[b]
\setlength{\tabcolsep}{14pt}
\centering
\caption{Cross Validation of MJ-COCO using Faster R-CNN \cite{ren2015faster}.}
\label{tab:faster_rcnn_comparison}
\resizebox{1.0\textwidth}{!}{%
\begin{tabular}{llcccccc}
\hline
Testing Data & Training Data & AP & AP$_{50}$ & AP$_{75}$ & AP$_S$ & AP$_M$ & AP$_L$ \\
\hline
\multirow{2}{*}{MS-COCO Validation \cite{lin2014microsoft}} 
& MS-COCO & \textbf{0.354} & \textbf{0.572} & \textbf{0.380} & \textbf{0.204} & \textbf{0.391} & \textbf{0.462} \\
& MJ-COCO   & 0.348 & 0.558 & 0.370 & 0.200 & 0.384 & 0.445 \\
\hline
\multirow{2}{*}{Sama Validation \cite{zimmermann2023benchmarking}} 
& MS-COCO & 0.343 & 0.550 & 0.365 & 0.196 & 0.384 & \textbf{0.473} \\
& MJ-COCO    & \textbf{0.354} & \textbf{0.562} & \textbf{0.375} & \textbf{0.212} & \textbf{0.395} & 0.470 \\
\hline
\multirow{2}{*}{Objects365 Validation \cite{shao2019objects365}} 
& MS-COCO  & 0.217 & 0.353 & 0.232 & 0.079 & 0.204 & 0.327 \\
& MJ-COCO     & \textbf{0.229} & \textbf{0.368} & \textbf{0.246} & \textbf{0.086} & \textbf{0.218} & \textbf{0.338} \\
\hline
\multirow{2}{*}{PASCAL Validation\cite{everingham2011pascal} } 
& MS-COCO  & 0.488 & \textbf{0.774} & 0.531 & \textbf{0.190} & 0.408 & 0.551 \\
& MJ-COCO   & \textbf{0.489} & \textbf{0.774} & \textbf{0.535} & 0.182 & \textbf{0.411} & \textbf{0.556} \\
\hline
\end{tabular}%
}
\end{table*}

\begin{table*}[t!]
\setlength{\tabcolsep}{14pt}
\centering
\caption{Cross Validation of MJ-COCO using Libra-RCNN\cite{pang2019libra}.}
\label{tab:libra_rcnn_comparison}
\resizebox{1.0\textwidth}{!}{%
\begin{tabular}{llcccccc}
\hline
Testing Data & Training Data & AP & AP$_{50}$ & AP$_{75}$ & AP$_S$ & AP$_M$ & AP$_L$ \\
\hline
\multirow{2}{*}{MS-COCO Validation \cite{lin2014microsoft}} 
& MS-COCO  & \textbf{0.360} & \textbf{0.576} & \textbf{0.387} & \textbf{0.207} & \textbf{0.400} & \textbf{0.464} \\
& MJ-COCO     & 0.355 & 0.565 & 0.382 & \textbf{0.207} & 0.391 & 0.452 \\
\hline
\multirow{2}{*}{Sama Validation \cite{zimmermann2023benchmarking}} 
& MS-COCO & 0.349 & 0.554 & 0.373 & 0.196 & 0.389 & 0.477 \\
& MJ-COCO  & \textbf{0.363} & \textbf{0.573} & \textbf{0.389} & \textbf{0.210} & \textbf{0.404} & \textbf{0.478} \\
\hline
\multirow{2}{*}{Objects365 Validation \cite{shao2019objects365}} 
& MS-COCO  & 0.221 & 0.355 & 0.240 & 0.082 & 0.207 & 0.333 \\
& MJ-COCO     & \textbf{0.232} & \textbf{0.368} & \textbf{0.253} & \textbf{0.089} & \textbf{0.219} & \textbf{0.344} \\
\hline
\multirow{2}{*}{PASCAL Validation\cite{everingham2011pascal} } 
& MS-COCO  & 0.491 & 0.770 & 0.538 & 0.180 & 0.413 & 0.560 \\
& MJ-COCO   & \textbf{0.496} & \textbf{0.773} & \textbf{0.546} & \textbf{0.194} & \textbf{0.421} & \textbf{0.569} \\
\hline
\end{tabular}%
}
\end{table*}

In Stage 3, bounding boxes with moderate-to-low confidence scores (below 0.6) underwent additional verification to further enhance annotation reliability. Specifically, cropped regions corresponding to these bounding boxes were evaluated using a separately trained ResNet-50 image classification model. Bounding boxes were retained or discarded based on a strict classification threshold criterion aligned with their confidence scores. For example, bounding boxes with confidence scores between 0.3 and 0.6 required a classification threshold probability of at least 0.7 to be retained, while boxes with lower confidence scores had even stricter thresholds. This rigorous verification procedure effectively eliminated incorrectly classified or ambiguous annotations, substantially improving dataset re-annotation. 
Finally, in Stage 4, we employed Grad-CAM to visually refine bounding boxes that passed previous verification stages. Grad-CAM provided activation maps highlighting the regions of the bounding boxes that contributed most significantly to the classification model's predictions. These activation maps were analyzed to assess the positional accuracy of bounding boxes. Bounding boxes displaying activation regions evenly centered on the object were retained without further modification. In contrast, bounding boxes with off-centered or inconsistent activation patterns were carefully adjusted or removed to align precisely with the object's actual location and boundaries. This Grad-CAM-based visual refinement was particularly effective in correcting subtle localization errors and semantic labeling inaccuracies, as visually demonstrated in the illustrative examples provided in Figure~\ref{PsudoLabelling_CAM}. The integration of Grad-CAM not only improved the localization accuracy but also ensured that annotations were semantically consistent with the classification model’s predictions, thus improving the reliability of pseudo labels.
To comprehensively evaluate improvements introduced by our pseudo-labeling approach, annotations were analyzed according to object size categories small, medium, and large, revealing substantial improvements across each category.
Small objects often face annotation inaccuracies due to their small footprint and frequent occlusions. Our re-annotation approach significantly improved annotations for several such categories. Notably, classes like Apple (+13,676), Carrot (+7,559), Bottle (+8,113), and Book (+10,997) showed large gains, indicating successful recovery of missed instances. Other classes such as Handbag (+2,170) and Wine Glass (+516) also benefited from refined instance separation. 
Minor reductions in classes like Fork (-295), Knife (-1,073), and Toothbrush (-53) reflect the effective removal of duplicates or incorrectly localized annotations, as shown in Table 5. These negative differences indicate the intentional removal of duplicated or inaccurately labeled instances.
Medium-sized objects frequently suffer from fragmented or overlapping annotations. Our method led to notable improvements in classes such as Banana (+40,247), Orange (+12,017), Chair (+18,259), and Donut (+4,443). These gains demonstrate improved instance consistency and boundary definition. Furthermore, objects like Traffic Light (+6,699) and Vase (+3,071) showed strong increases due to better differentiation in crowded scenes. Slight declines in categories like Bowl (-767) and Snowboard (-120) suggest that the method also eliminated redundant or poorly annotated instances as given in Table 6.
Large-scale objects typically present challenges with localization precision and over-annotation, especially in complex scenes with dense object arrangements. Our pseudo-labeling approach led to measurable improvements in classes such as Airplane (+675), Bus (+1,063), Teddy Bear (+1,639), and Truck (+1,503), reflecting enhanced bounding box quality and reduced label noise. Additionally, increased counts in Train (+312) and Pizza (+228) support improved coverage of partial or occluded instances. Reductions in classes such as Bed (-15), Couch (-181), and Sandwich (-448) highlight successful removal of incorrect or redundant annotations as given in Table 7.
The effectiveness of the proposed pseudo-labeling method is visually shown in Figure 5, which presents a side-by-side comparison of object annotations from the original MS-COCO dataset (highlighted in red) and the corrected annotations produced by our method (highlighted in green).
\textcolor{black}{A detailed breakdown of the class-wise AP performance is given in supplementary file Table 4}.


\subsection{Comparison with State-of-the-Art Methods}
To rigorously evaluate the effectiveness and generalization capability of the proposed pseudo-labeling strategy, comprehensive comparisons were conducted against state-of-the-art OD models trained on the newly constructed MJ-COCO dataset. Specifically, we benchmarked widely used one-stage models including RetinaNet\cite{lin2017focal}, YOLOv3\cite{redmon2018yolov3}, YOLOX \cite{ge2021yolox} and two-stage models such as Faster R-CNN\cite{ren2015faster}, Libra R-CNN \cite{pang2019libra}. Performance was quantitatively assessed using multiple evaluation metrics, including AP at fixed IoU thresholds (AP$_{50}$, AP$_{75}$), and size-specific AP metrics (AP$_S$, AP$_M$, AP$_L$). To ensure unbiased and reproducible results, all models were trained and evaluated under identical conditions utilizing the standardized MMDetection framework. Performance evaluations were conducted using four separate validation datasets: the original MS-COCO validation set, Sama COCO validation set \cite{zimmermann2023benchmarking}, Objects365\cite{shao2019objects365}, and PASCAL\cite{everingham2011pascal} datasets. This comprehensive evaluation enabled a clear assessment of our MJ-COCO dataset's impact on model performance, particularly in realistic and rigorously annotated scenarios.
\subsubsection{\textbf{Performance of One-Stage Object Detection Models}}
In the OD domain, single-stage detection models are highly significant due to their computational efficiency, real-time inference capabilities, and applications. To comparatively analyze their performance, we evaluated prominent single-stage models such as RetinaNet, YOLOv3, and YOLOX, using validation sets from standard benchmark datasets such as the original MS-COCO, Sama, Objects365, and PASCAL. Based on these comprehensive evaluations, we effectively assessed the generalization capability and robustness of our proposed MJ-COCO dataset across. Detailed comparative results are given in Tables~\ref{tab:retinanet_comparison},~\ref{tab:yolov3_comparison}, and~\ref{tab:yolox_comparison}.
In the case of RetinaNet \textcolor{black}{results as given in Table~\ref{tab:retinanet_comparison}}, the MJ-COCO dataset consistently enhanced model performance in rigorously annotated external validation scenarios. Specifically, improvements were observed on the Sama validation set with increases in AP (from 0.356 to 0.358), AP$_{50}$ (from 0.532 to 0.538), and notably AP$_S$ (from 0.198 to 0.208). Similarly, improvements on the Objects365 validation set suggest MJ-COCO’s annotations significantly enhance detection performance for challenging classes. For YOLOv3, Table~\ref{tab:yolov3_comparison} shows that models trained on MJ-COCO outperformed the original MS-COCO trained counterparts on the Sama COCO validation set, showing improvements in AP (from 0.281 to 0.286) and AP$_{50}$ (from 0.484 to 0.499). Similar gains were apparent on the Objects365 dataset, further validating the robustness and effectiveness of the MJ-COCO annotations across diverse object categories. YOLOX results, as given in Table~\ref{tab:yolox_comparison},  exhibited significant gains with MJ-COCO on Sama validation, improving AP (from 0.447 to 0.457), AP$_{50}$ (from 0.618 to 0.632), and AP$_{75}$ (from 0.483 to 0.493). Crucially, substantial performance improvements were also observed on Objects365 and minor gains on PASCAL, further emphasizing the generalization capability of MJ-COCO annotations.

\subsubsection{\textbf{Performance of Two-Stage Object Detection Model}}

Two-stage OD models are extensively employed in computer vision due to their superior accuracy in precise localization tasks, especially under challenging conditions. To rigorously evaluate their performance, we benchmarked prominent two-stage detectors, specifically Faster R-CNN and Libra R-CNN. Comprehensive comparative results, highlighting the improved generalization provided by our proposed MJ-COCO dataset are given in Table~\ref{tab:faster_rcnn_comparison} and ~\ref{tab:libra_rcnn_comparison}.

For Faster R-CNN (Table~\ref{tab:faster_rcnn_comparison}), the MJ-COCO dataset consistently improved performance metrics, particularly notable on the Sama validation set with AP improvements (from 0.343 to 0.354), AP$_{50}$ (from 0.550 to 0.562), and AP$_S$ (from 0.196 to 0.212). Additionally, consistent gains on the Objects365 dataset further confirmed the suitability and precision of the MJ-COCO annotations. Libra R-CNN (Table~\ref{tab:libra_rcnn_comparison}) also demonstrated performance enhancements on the Sama validation set with notable increases in AP (from 0.349 to 0.363) and AP$_{50}$ (from 0.554 to 0.573). Furthermore, consistent improvements observed on the Objects365 and PASCAL benchmarks strongly indicate that MJ-COCO effectively addresses annotation inaccuracies, significantly benefiting localization-sensitive detection tasks. The extensive evaluation clearly demonstrates the substantial impact and importance of our proposed MJ-COCO dataset. Specifically, MJ-COCO achieved consistent improvements across both one-stage and two-stage state-of-the-art OD models, with performance gains notably visible in rigorously annotated validation datasets such as Sama COCO and Objects365. The improved AP, AP$_{50}$, AP$_{75}$, and small-OD metrics (AP$_S$) highlight MJ-COCO’s superior annotation quality and precision. Thus, MJ-COCO provides a reliable and robust benchmark, significantly benefiting the OD research community by addressing critical annotation challenges present in existing datasets.

\section{Conclusions} 

In this study, we addressed the critical issue of annotation errors in large-scale object detection datasets, with a focus on the widely used MS-COCO dataset. We proposed a comprehensive pipeline that first detects anomalous samples during model training using loss and gradient-based anomaly detection, followed by a robust pseudo-labeling strategy enhanced with Grad-CAM verification, invertible transformation consistency, and confidence-based filtering. This methodology enabled the creation of a refined dataset, MJ-COCO, which corrects and supplements flawed annotations in the original dataset.

To evaluate the quality and effectiveness of MJ-COCO, we conducted extensive experiments across a range of state-of-the-art one-stage and two-stage object detection models, including RetinaNet, YOLOv3, YOLOX, Faster R-CNN, and Libra R-CNN. Validation was performed on multiple benchmark datasets MS-COCO, Sama, Objects365, and PASCAL demonstrating that models trained on MJ-COCO consistently outperformed or matched their MS-COCO counterparts, especially on rigorously annotated benchmarks. Notable improvements were observed in AP, especially AP$_{50}$ and AP$_S$, highlighting the benefit of high-quality labels for small and challenging objects.
The MJ-COCO dataset significantly enhances the reliability and generalizability of object detection models by mitigating the impact of annotation noise. Our results indicate that careful re-labeling using guided pseudo-labeling and anomaly detection not only improves model accuracy but also provides a scalable framework for refining existing datasets. This work contributes a practical and effective approach for improving dataset quality, which is fundamental for advancing research in object detection and related areas.

\label{sec:conclusion}

\section*{Data Availability Statement}
The newly created dataset, named MJ-COCO 2025, is publicly available and can be accessed via the link: \url{https://www.kaggle.com/datasets/mjcoco2025/mj-coco-2025}


\begin{thebibliography}{10}
\providecommand{\url}[1]{#1}
\csname url@samestyle\endcsname
\providecommand{\newblock}{\relax}
\providecommand{\bibinfo}[2]{#2}
\providecommand{\BIBentrySTDinterwordspacing}{\spaceskip=0pt\relax}
\providecommand{\BIBentryALTinterwordstretchfactor}{4}
\providecommand{\BIBentryALTinterwordspacing}{\spaceskip=\fontdimen2\font plus
\BIBentryALTinterwordstretchfactor\fontdimen3\font minus \fontdimen4\font\relax}
\providecommand{\BIBforeignlanguage}[2]{{%
\expandafter\ifx\csname l@#1\endcsname\relax
\typeout{** WARNING: IEEEtran.bst: No hyphenation pattern has been}%
\typeout{** loaded for the language `#1'. Using the pattern for}%
\typeout{** the default language instead.}%
\else
\language=\csname l@#1\endcsname
\fi
#2}}
\providecommand{\BIBdecl}{\relax}
\BIBdecl

\bibitem{bolya2020tide}
D.~Bolya, S.~Foley, J.~Hays, and J.~Hoffman, ``Tide: A general toolbox for identifying object detection errors,'' in \emph{Computer Vision--ECCV 2020: 16th European Conference, Glasgow, UK, August 23--28, 2020, Proceedings, Part III 16}.\hskip 1em plus 0.5em minus 0.4em\relax Springer, 2020, pp. 558--573.

\bibitem{papageorgiou1998trainable}
C.~Papageorgiou, T.~Evgeniou, and T.~Poggio, ``A trainable pedestrian detection system,'' in \emph{Proc. of Intelligent Vehicles}, 1998, pp. 241--246.

\bibitem{dalal2005histograms}
N.~Dalal and B.~Triggs, ``Histograms of oriented gradients for human detection,'' in \emph{2005 IEEE computer society conference on computer vision and pattern recognition (CVPR'05)}, vol.~1.\hskip 1em plus 0.5em minus 0.4em\relax Ieee, 2005, pp. 886--893.

\bibitem{everingham20062005}
M.~Everingham, A.~Zisserman, C.~K. Williams, L.~Van~Gool, M.~Allan, C.~M. Bishop, O.~Chapelle, N.~Dalal, T.~Deselaers, G.~Dork{\'o} \emph{et~al.}, ``The 2005 pascal visual object classes challenge,'' in \emph{Machine learning challenges. evaluating predictive uncertainty, visual object classification, and recognising tectual entailment: first PASCAL machine learning challenges workshop, MLCW 2005, southampton, UK, April 11-13, 2005, revised selected papers}.\hskip 1em plus 0.5em minus 0.4em\relax Springer, 2006, pp. 117--176.

\bibitem{everingham2011pascal}
M.~Everingham and J.~Winn, ``The pascal visual object classes challenge 2012 (voc2012) development kit,'' \emph{Pattern Analysis, Statistical Modelling and Computational Learning, Tech. Rep}, vol.~8, no.~5, pp. 2--5, 2011.

\bibitem{mscoco}
{C. Consortium}, ``{MS COCO: Common Objects in Context},'' \url{https://cocodataset.org/#download}, 2014, accessed: 2025-04-07.

\bibitem{zimmermann2023benchmarking}
E.~Zimmermann, J.~Szeto, J.~Pasquero, and F.~Ratle, ``Benchmarking a benchmark: How reliable is ms-coco?'' \emph{arXiv preprint arXiv:2311.02709}, 2023.

\bibitem{ma2022effect}
J.~Ma, Y.~Ushiku, and M.~Sagara, ``The effect of improving annotation quality on object detection datasets: A preliminary study,'' in \emph{Proceedings Of The IEEE/CVF conference on computer vision and pattern recognition}, 2022, pp. 4850--4859.

\bibitem{everingham2006pascal}
M.~Everingham, A.~Zisserman, C.~Williams, L.~Van~Gool \emph{et~al.}, ``The pascal visual object classes challenge 2006 (voc 2006) results (technical report). september 2006,'' \emph{The PASCAL2006 dataset can be downloaded at http://www. pascal-network. org/challenges/VOC/voc2006}, 2006.

\bibitem{everingham2008pascal}
M.~Everingham, L.~Van~Gool, C.~K. Williams, J.~Winn, and A.~Zisserman, ``The pascal visual object classes challenge 2007 (voc 2007) results (2007),'' 2008.

\bibitem{hoiem2009pascal}
D.~Hoiem, S.~K. Divvala, and J.~H. Hays, ``Pascal voc 2008 challenge,'' \emph{World Literature Today}, vol.~24, no.~1, pp. 1--4, 2009.

\bibitem{everingham2009pascal}
M.~Everingham, L.~VanGool, C.~Williams, J.~Winn, and A.~Zisserman, ``The pascal visual object classes challenge 2009 (voc2009) results. http,'' in \emph{www. pascal-network. org/challenges/VOC/voc2009/workshop/index. html}, 2009.

\bibitem{everingham2010pascal}
M.~Everingham, L.~Van~Gool, C.~K. Williams, J.~Winn, and A.~Zisserman, ``The pascal visual object classes (voc) challenge,'' \emph{International journal of computer vision}, vol.~88, pp. 303--338, 2010.

\bibitem{everinghampascal}
M.~Everingham, L.~Van~Gool, C.~Williams, J.~Winn, and A.~Zisserman, ``The pascal visual object classes challenge 2011 (voc2011) results, 2011< http://www. pascal-network. org/challenges,'' in \emph{VOC/voc2011/workshop/index. html}.

\bibitem{oksuz2020imbalance}
K.~Oksuz, B.~C. Cam, S.~Kalkan, and E.~Akbas, ``Imbalance problems in object detection: A review,'' \emph{IEEE transactions on pattern analysis and machine intelligence}, vol.~43, no.~10, pp. 3388--3415, 2020.

\bibitem{borjibreaking}
A.~Borji, ``Breaking beyond coco object detection.''

\bibitem{tong2023rethinking}
K.~Tong and Y.~Wu, ``Rethinking pascal-voc and ms-coco dataset for small object detection,'' \emph{Journal of Visual Communication and Image Representation}, vol.~93, p. 103830, 2023.

\bibitem{zhu2015diagnosing}
H.~Zhu, S.~Lu, J.~Cai, and Q.~Lee, ``Diagnosing state-of-the-art object proposal methods,'' \emph{arXiv preprint arXiv:1507.04512}, 2015.

\bibitem{murrugarra2022can}
J.~Murrugarra-Llerena, L.~N. Kirsten, and C.~R. Jung, ``Can we trust bounding box annotations for object detection?'' in \emph{Proceedings of the IEEE/CVF Conference on Computer Vision and Pattern Recognition}, 2022, pp. 4813--4822.

\bibitem{tkachenko2023objectlab}
U.~Tkachenko, A.~Thyagarajan, and J.~Mueller, ``Objectlab: Automated diagnosis of mislabeled images in object detection data,'' \emph{arXiv preprint arXiv:2309.00832}, 2023.

\bibitem{schubert2024identifying}
M.~Schubert, T.~Riedlinger, K.~Kahl, D.~Kr{\"o}ll, S.~Schoenen, S.~{\v{S}}egvi{\'c}, and M.~Rottmann, ``Identifying label errors in object detection datasets by loss inspection,'' in \emph{Proceedings of the IEEE/CVF Winter Conference on Applications of Computer Vision}, 2024, pp. 4582--4591.

\bibitem{xu2021training}
Y.~Xu, L.~Zhu, Y.~Yang, and F.~Wu, ``Training robust object detectors from noisy category labels and imprecise bounding boxes,'' \emph{IEEE Transactions on Image Processing}, vol.~30, pp. 5782--5792, 2021.

\bibitem{liu2022robust}
C.~Liu, K.~Wang, H.~Lu, Z.~Cao, and Z.~Zhang, ``Robust object detection with inaccurate bounding boxes,'' in \emph{European Conference on Computer Vision}.\hskip 1em plus 0.5em minus 0.4em\relax Springer, 2022, pp. 53--69.

\bibitem{li2022pseco}
G.~Li, X.~Li, Y.~Wang, Y.~Wu, D.~Liang, and S.~Zhang, ``Pseco: Pseudo labeling and consistency training for semi-supervised object detection,'' in \emph{European Conference on Computer Vision}.\hskip 1em plus 0.5em minus 0.4em\relax Springer, 2022, pp. 457--472.

\bibitem{zhou2022dense}
H.~Zhou, Z.~Ge, S.~Liu, W.~Mao, Z.~Li, H.~Yu, and J.~Sun, ``Dense teacher: Dense pseudo-labels for semi-supervised object detection,'' in \emph{European Conference on Computer Vision}.\hskip 1em plus 0.5em minus 0.4em\relax Springer, 2022, pp. 35--50.

\bibitem{yang2025pseudo}
K.~Yang, Y.~Wu, J.~Li, C.~Yin, and X.~Li, ``Pseudo-label enhancement for weakly supervised object detection using self-supervised vision transformer,'' \emph{Knowledge-Based Systems}, vol. 311, p. 113012, 2025.

\bibitem{garcia2025enhancing}
P.~Garcia-Fernandez, D.~Cores, and M.~Mucientes, ``Enhancing few-shot object detection through pseudo-label mining,'' \emph{Image and Vision Computing}, vol. 154, p. 105379, 2025.

\bibitem{ren2016faster}
S.~Ren, K.~He, R.~Girshick, and J.~Sun, ``Faster r-cnn: Towards real-time object detection with region proposal networks,'' \emph{IEEE transactions on pattern analysis and machine intelligence}, vol.~39, no.~6, pp. 1137--1149, 2016.

\bibitem{lin2014microsoft}
T.-Y. Lin, M.~Maire, S.~Belongie, J.~Hays, P.~Perona, D.~Ramanan, P.~Doll{\'a}r, and C.~L. Zitnick, ``Microsoft coco: Common objects in context,'' in \emph{Computer vision--ECCV 2014: 13th European conference, zurich, Switzerland, September 6-12, 2014, proceedings, part v 13}.\hskip 1em plus 0.5em minus 0.4em\relax Springer, 2014, pp. 740--755.

\bibitem{ge2021yolox}
Z.~Ge, S.~Liu, F.~Wang, Z.~Li, and J.~Sun, ``Yolox: Exceeding yolo series in 2021,'' \emph{arXiv preprint arXiv:2107.08430}, 2021.

\bibitem{lin2017focal}
T.-Y. Lin, P.~Goyal, R.~Girshick, K.~He, and P.~Doll{\'a}r, ``Focal loss for dense object detection,'' in \emph{Proceedings of the IEEE international conference on computer vision}, 2017, pp. 2980--2988.

\bibitem{shao2019objects365}
S.~Shao, Z.~Li, T.~Zhang, C.~Peng, G.~Yu, X.~Zhang, J.~Li, and J.~Sun, ``Objects365: A large-scale, high-quality dataset for object detection,'' in \emph{Proceedings of the IEEE/CVF international conference on computer vision}, 2019, pp. 8430--8439.

\bibitem{griffin2007caltech}
G.~Griffin, A.~Holub, P.~Perona \emph{et~al.}, ``Caltech-256 object category dataset,'' Technical Report 7694, California Institute of Technology Pasadena, Tech. Rep., 2007.

\bibitem{deng2009imagenet}
J.~Deng, W.~Dong, R.~Socher, L.-J. Li, K.~Li, and L.~Fei-Fei, ``Imagenet: A large-scale hierarchical image database,'' in \emph{2009 IEEE conference on computer vision and pattern recognition}.\hskip 1em plus 0.5em minus 0.4em\relax Ieee, 2009, pp. 248--255.

\bibitem{krizhevsky2010cifar}
A.~Krizhevsky, V.~Nair, and G.~Hinton, ``Cifar-10 (canadian institute for advanced research),'' \emph{URL http://www. cs. toronto. edu/kriz/cifar. html}, vol.~5, no.~4, p.~1, 2010.

\bibitem{ciaglia2022roboflow}
F.~Ciaglia, F.~S. Zuppichini, P.~Guerrie, M.~McQuade, and J.~Solawetz, ``Roboflow 100: A rich, multi-domain object detection benchmark,'' \emph{arXiv preprint arXiv:2211.13523}, 2022.

\bibitem{redmon2018yolov3}
J.~Redmon and A.~Farhadi, ``Yolov3: An incremental improvement,'' \emph{arXiv preprint arXiv:1804.02767}, 2018.

\bibitem{ren2015faster}
S.~Ren, ``Faster r-cnn: Towards real-time object detection with region proposal networks,'' \emph{arXiv preprint arXiv:1506.01497}, 2015.

\bibitem{pang2019libra}
J.~Pang, K.~Chen, J.~Shi, H.~Feng, W.~Ouyang, and D.~Lin, ``Libra r-cnn: Towards balanced learning for object detection,'' in \emph{Proceedings of the IEEE/CVF conference on computer vision and pattern recognition}, 2019, pp. 821--830.

\end{thebibliography}

\end{document}


\title{Supplementary Material for:\\
Pseudo-Labeling Driven Refinement of Benchmark
Object Detection Datasets via Analysis of Learning
Patterns}

\author{Min~Je~Kim, Muhammad Munsif,~\IEEEmembership{Student Member,~IEEE}, Altaf Hussain,~\IEEEmembership{Student Member,~IEEE}, Hikmat Yar,~\IEEEmembership{Student Member,~IEEE},
              Sung Wook~Baik,~\IEEEmembership{Senior Member,~IEEE}

}

\maketitle

\begin{table}[h]
\setlength{\tabcolsep}{10pt}
\centering
\caption{AUROC and PRAUC Performance According to Loss and Gradient Weight Variation (AutoEncoder).}
\resizebox{0.9\textwidth}{!}{
\label{tab:auroc_prauc_lambdaOnly}
\begin{tabular}{l|ccccccccc}
\hline
\textbf{$\lambda$} & 0.1 & 0.2 & 0.3 & 0.4 & 0.5 & 0.6 & 0.7 & 0.8 & 0.9 \\
\hline
\textbf{AUROC} & 0.7931 & 0.8171 & 0.8302 & 0.8383 & 0.8436 & 0.8474 & 0.8500 & 0.8520 & 0.8534 \\
\textbf{PRAUC} & 0.6130 & 0.6538 & 0.6804 & 0.6985 & 0.7113 & 0.7209 & 0.7281 & 0.7337 & 0.7378 \\
\hline
\end{tabular}}
\end{table}

\begin{table}[H]
\setlength{\tabcolsep}{10pt}
\centering
\caption{Performance Evaluation According to Feature-Specific Weight Changes.}
\resizebox{0.9\textwidth}{!}{
\begin{tabular}{cccccccccc}
\hline
$\alpha$ & $\beta$ & $\gamma$ & $\delta$ & Accuracy & Precision & Recall & F1-score & AUROC & PRAUC \\
\hline
0.1 & 0.1 & 0.1 & 0.7 & 0.7873 & 0.6938 & 0.6971 & 0.6955 & 0.8373 & 0.7132 \\
0.1 & 0.1 & 0.3 & 0.5 & 0.7898 & 0.6973 & 0.7007 & 0.6990 & 0.8416 & 0.7150 \\
0.1 & 0.1 & 0.5 & 0.3 & 0.7931 & 0.7020 & 0.7054 & 0.7037 & 0.8459 & 0.7144 \\
0.1 & 0.1 & 0.7 & 0.1 & 0.7908 & 0.6988 & 0.7021 & 0.7004 & 0.8461 & 0.7020 \\
0.1 & 0.3 & 0.1 & 0.5 & 0.7874 & 0.6939 & 0.6973 & 0.6956 & 0.8383 & 0.7106 \\
0.1 & 0.3 & 0.3 & 0.3 & 0.7894 & 0.6968 & 0.7002 & 0.6985 & 0.8420 & 0.7079 \\
0.1 & 0.3 & 0.5 & 0.1 & 0.7863 & 0.6923 & 0.6956 & 0.6939 & 0.8405 & 0.6910 \\
0.1 & 0.5 & 0.1 & 0.3 & 0.7845 & 0.6897 & 0.6931 & 0.6914 & 0.8349 & 0.6957 \\
0.1 & 0.5 & 0.3 & 0.1 & 0.7781 & 0.6807 & 0.6840 & 0.6823 & 0.8307 & 0.6720 \\
0.1 & 0.7 & 0.1 & 0.1 & 0.7670 & 0.6647 & 0.6679 & 0.6663 & 0.8173 & 0.6493 \\
0.3 & 0.1 & 0.1 & 0.5 & 0.7904 & 0.6982 & 0.7016 & 0.6999 & 0.8409 & 0.7177 \\
\textbf{0.3} & \textbf{0.1} & \textbf{0.3} & \textbf{0.3} & \textbf{0.7939} & \textbf{0.7031} & \textbf{0.7065} & \textbf{0.7048} & \textbf{0.8459} & \textbf{0.7190} \\
0.3 & 0.1 & 0.5 & 0.1 & 0.7918 & 0.7002 & 0.7036 & 0.7019 & 0.8471 & 0.7089 \\
0.3 & 0.3 & 0.1 & 0.3 & 0.7899 & 0.6975 & 0.7009 & 0.6992 & 0.8405 & 0.7102 \\
0.3 & 0.3 & 0.3 & 0.1 & 0.7862 & 0.6921 & 0.6955 & 0.6938 & 0.8392 & 0.6940 \\
0.3 & 0.5 & 0.1 & 0.1 & 0.7765 & 0.6783 & 0.6816 & 0.6799 & 0.8260 & 0.6703 \\
0.5 & 0.1 & 0.1 & 0.3 & 0.7924 & 0.7011 & 0.7045 & 0.7028 & 0.8429 & 0.7198 \\
0.5 & 0.1 & 0.3 & 0.1 & 0.7907 & 0.6986 & 0.7020 & 0.7003 & 0.8434 & 0.7097 \\
0.5 & 0.3 & 0.1 & 0.1 & 0.7826 & 0.6870 & 0.6903 & 0.6887 & 0.8318 & 0.6896 \\
0.7 & 0.1 & 0.1 & 0.1 & 0.7850 & 0.6904 & 0.6937 & 0.6920 & 0.8329 & 0.7025 \\
\hline
\end{tabular}}
\label{tab:feature_specific_weights}
\end{table}

\begin{table}[H]
\setlength{\tabcolsep}{10pt}
\centering
\caption{Comparative Analysis of the Proposed AutoEncoder-Based Anomaly Detection with and without Weighted.}
\resizebox{0.9\textwidth}{!}{
\label{tab:finalWeightsComparisionAuto}
\begin{tabular}{lcccccc}
\hline
\textbf{Method} & \textbf{Acc} & \textbf{Prec} & \textbf{Recall} & \textbf{F1} & \textbf{AUROC} & \textbf{PRAUC} \\
\hline
No Weighting Applied & 0.7916 & 0.6999 & 0.7033 & 0.7016 & 0.8436 & 0.7114 \\
Optimal  $\lambda$ Applied & 0.7982 & 0.7093 & 0.7127 & 0.7110 & 0.8534 & 0.7378 \\
Optimal $\alpha$-$\delta$ Applied & 0.7939 & 0.7031 & 0.7065 & 0.7048 & 0.8459 & 0.7190 \\
\textbf{Combined Weighting Applied} & \textbf{0.8008} & \textbf{0.7131} & \textbf{0.7165} & \textbf{0.7148} & \textbf{0.8572} & \textbf{0.7523} \\
\hline
\end{tabular}}
\end{table}

\begin{table}[htbp]
\setlength{\tabcolsep}{5pt}
\centering
\caption{Performance of Faster-RCNN for Each Class using Average Precision.}
\resizebox{0.9\textwidth}{!}{\label{ap_change_by_class_8col}
\scriptsize
\begin{tabular}{lccclccc}
\toprule
Class & MS-COCO & MJ-COCO & Different & Class & MS-COCO & MJ-COCO & Different \\
\midrule
Airplane & 0.539 & 0.555 & 0.016 & Apple & 0.135 & 0.251 & 0.116 \\
Backpack & 0.131 & 0.121 & -0.010 & Banana & 0.059 & 0.197 & 0.138 \\
Baseball bat & 0.163 & 0.137 & -0.026 & Baseball Glove & 0.369 & 0.341 & -0.028 \\
Bear & 0.571 & 0.584 & 0.013 & Bed & 0.376 & 0.374 & -0.002 \\
Bench & 0.194 & 0.197 & 0.003 & Bicycle & 0.281 & 0.278 & -0.003 \\
Bird & 0.313 & 0.334 & 0.021 & Boat & 0.238 & 0.240 & 0.002 \\
Book & 0.145 & 0.140 & -0.005 & Bottle & 0.323 & 0.327 & 0.004 \\
Bowl & 0.369 & 0.379 & 0.010 & Broccoli & 0.204 & 0.267 & 0.063 \\
Bus & 0.541 & 0.557 & 0.016 & Cake & 0.257 & 0.277 & 0.020 \\
Car & 0.382 & 0.390 & 0.008 & Carrot & 0.101 & 0.153 & 0.052 \\
Cat & 0.539 & 0.567 & 0.028 & Cell Phone & 0.313 & 0.318 & 0.005 \\
Chair & 0.177 & 0.179 & 0.002 & Clock & 0.476 & 0.485 & 0.009 \\
Couch & 0.297 & 0.295 & -0.002 & Cow & 0.470 & 0.484 & 0.014 \\
Cup & 0.366 & 0.375 & 0.009 & Dining Table & 0.226 & 0.242 & 0.016 \\
Dog & 0.542 & 0.527 & -0.015 & Donut & 0.355 & 0.373 & 0.018 \\
Elephant & 0.544 & 0.570 & 0.026 & Fire Hydrant & 0.604 & 0.604 & 0.000 \\
Fork & 0.235 & 0.226 & -0.009 & Frisbee & 0.604 & 0.605 & 0.001 \\
Giraffe & 0.616 & 0.614 & -0.002 & Hair Drier & 0.024 & 0.003 & -0.021 \\
Handbag & 0.114 & 0.111 & -0.003 & Horse & 0.478 & 0.487 & 0.009 \\
Hot Dog & 0.310 & 0.335 & 0.025 & Keyboard & 0.395 & 0.422 & 0.027 \\
Kite & 0.331 & 0.353 & 0.022 & Knife & 0.121 & 0.127 & 0.006 \\
Laptop & 0.511 & 0.502 & -0.009 & Microwave & 0.495 & 0.502 & 0.007 \\
Motorcycle & 0.322 & 0.319 & -0.003 & Mouse & 0.588 & 0.584 & -0.004 \\
Orange & 0.340 & 0.409 & 0.069 & Oven & 0.391 & 0.317 & -0.074 \\
Parking Meter & 0.466 & 0.462 & -0.004 & Person & 0.408 & 0.459 & 0.051 \\
Pizza & 0.495 & 0.490 & -0.005 & Potted Plant & 0.112 & 0.128 & 0.016 \\
Refrigerator & 0.447 & 0.461 & 0.014 & Remote & 0.221 & 0.220 & -0.001 \\
Sandwich & 0.267 & 0.286 & 0.019 & Scissors & 0.225 & 0.231 & 0.006 \\
Sheep & 0.443 & 0.481 & 0.038 & Sink & 0.319 & 0.320 & 0.001 \\
Skateboard & 0.423 & 0.424 & 0.001 & Skis & 0.097 & 0.121 & 0.024 \\
Snowboard & 0.293 & 0.306 & 0.013 & Spoon & 0.094 & 0.099 & 0.005 \\
Sports Ball & 0.475 & 0.475 & 0.000 & Stop Sign & 0.717 & 0.700 & -0.017 \\
Suitcase & 0.218 & 0.227 & 0.009 & Surfboard & 0.288 & 0.301 & 0.013 \\
Teddy Bear & 0.371 & 0.390 & 0.019 & Tennis Racket & 0.415 & 0.396 & -0.019 \\
Tie & 0.229 & 0.230 & 0.001 & Toaster & 0.335 & 0.371 & 0.036 \\
Toilet & 0.518 & 0.524 & 0.006 & Toothbrush & 0.219 & 0.184 & -0.035 \\
Traffic Light & 0.305 & 0.304 & -0.001 & Train & 0.507 & 0.529 & 0.022 \\
Truck & 0.242 & 0.235 & -0.007 & TV & 0.483 & 0.498 & 0.015 \\
Umbrella & 0.236 & 0.258 & 0.022 & Vase & 0.227 & 0.278 & 0.051 \\
Wine Glass & 0.292 & 0.307 & 0.015 & Zebra & 0.564 & 0.577 & 0.013 \\

\bottomrule
\end{tabular}}
\end{table}